\documentclass{article}

\usepackage{microtype}
\usepackage{graphicx}
\usepackage{booktabs} 
\usepackage{enumitem}
\usepackage{blindtext} 

\usepackage{amsthm}
\usepackage{bbm} 

\makeatletter
\newcommand*{\transpose}{%
  {\mathpalette\@transpose{}}%
}
\newcommand*{\@transpose}[2]{%
  \raisebox{\depth}{$\m@th#1\intercal$}%
}

\newcommand{\defeq}{\vcentcolon=}

\usepackage{amsfonts} 
\usepackage{amsmath} 
\usepackage{amssymb} 
\usepackage{mathtools} 
\usepackage{subcaption} 
\usepackage{wrapfig} 
\usepackage{nicefrac} 
\usepackage{multirow} 
\usepackage{arydshln} 

\usepackage{amsmath,amsfonts,bm}

\def\eqref#1{equation~\ref{#1}}

\def\1{\bm{1}}

\DeclareMathAlphabet{\mathsfit}{\encodingdefault}{\sfdefault}{m}{sl}
\SetMathAlphabet{\mathsfit}{bold}{\encodingdefault}{\sfdefault}{bx}{n}

\DeclareMathOperator*{\argmin}{arg\,min}

\usepackage{graphicx} 
\usepackage[export]{adjustbox}

\usepackage{xcolor}

\usepackage{algorithm} 
\usepackage{algpseudocode}
\usepackage{hyperref}
\usepackage{url}
\usepackage{adjustbox}

\renewcommand*{\thesection}{\Roman{section}}

\algblock{ParFor}{EndParFor}
\algnewcommand\algorithmicparfor{\textbf{parfor}}
\algnewcommand\algorithmicpardo{\textbf{do}}
\algnewcommand\algorithmicendparfor{\textbf{end\ parfor}}
\algrenewtext{ParFor}[1]{\algorithmicparfor\ #1\ \algorithmicpardo}
\algrenewtext{EndParFor}{\algorithmicendparfor}

\usepackage{booktabs}

\newenvironment{thmbis}[1]
  {\renewcommand{\thetheorem}{\ref{#1}$'$}%
   \addtocounter{theorem}{-1}%
   \begin{theorem}}
  {\end{theorem}}

\usepackage{natbib} 
\PassOptionsToPackage{sort}{natbib}
\definecolor{darkgreen}{rgb}{0.0, 0.5, 0.13}

\usepackage[accepted]{icml2024}

\usepackage{amsmath}
\usepackage{amssymb}
\usepackage{mathtools}
\usepackage{amsthm}

\usepackage{float}

\usepackage[capitalize,noabbrev,nameinlink]{cleveref}

\theoremstyle{plain}
\newtheorem{theorem}{Theorem}[section]

\newtheorem{lemma}[theorem]{Lemma}
\newtheorem{corollary}[theorem]{Corollary}
\theoremstyle{definition}

\theoremstyle{remark}
\newtheorem{remark}[theorem]{Remark}

\usepackage[textsize=tiny]{todonotes}

\icmltitlerunning{Embarrassingly Parallel GFlowNets}

\begin{document}

\twocolumn[
\icmltitle{Embarrassingly Parallel GFlowNets}



\icmlsetsymbol{equal}{*}

\begin{icmlauthorlist}
\icmlauthor{Tiago da Silva}{1}
\icmlauthor{Luiz Max Carvalho}{1}
\icmlauthor{Amauri Souza}{2,3}
\icmlauthor{Samuel Kaski}{2,4}
\icmlauthor{Diego Mesquita}{1}
\end{icmlauthorlist}



\icmlaffiliation{1}{Getulio Vargas Foundation}
\icmlaffiliation{2}{Aalto University}
\icmlaffiliation{3}{Federal Institute of Cear\'a}
\icmlaffiliation{4}{University of Manchester}

\icmlcorrespondingauthor{Diego Mesquita}{diego.mesquita@fgv.br}

\icmlkeywords{Machine Learning, ICML}

\vskip 0.3in
]



\printAffiliationsAndNotice{} 

\begin{abstract}
GFlowNets are a promising alternative to MCMC sampling for discrete compositional random variables. Training GFlowNets requires repeated evaluations of the unnormalized target distribution, or reward function. However, for large-scale posterior sampling, this may be prohibitive since it incurs traversing the data several times.
Moreover, if the data are distributed across clients, employing standard GFlowNets leads to intensive client-server communication.
To alleviate both these issues, we propose \emph{embarrassingly parallel} GFlowNet (EP-GFlowNet). EP-GFlowNet is a provably correct divide-and-conquer method to sample from product distributions of the form $R(\cdot) \propto R_1(\cdot) ... R_N(\cdot)$ --- e.g., in parallel or federated Bayes, where each $R_n$ is a local posterior defined on a data partition. 
First, in parallel, we train a local GFlowNet targeting each $R_n$ and send the resulting models to the server. Then, the server learns a global GFlowNet by enforcing our newly proposed \emph{aggregating balance} condition, requiring a single communication step.
Importantly, EP-GFlowNets can also be applied to multi-objective optimization and model reuse.
Our experiments illustrate the EP-GFlowNets's effectiveness on many tasks, including parallel Bayesian phylogenetics, multi-objective multiset, 
 sequence generation, and federated Bayesian structure learning. 
\end{abstract}


\section{Introduction}

Generative Flow Networks (GFlowNets) \citep{Bengio2021} are powerful sampling tools and, despite being in their infancy, have become popular alternatives to Markov Chain Monte Carlo methods in finite discrete state spaces. While the most straight-forward use of MCMC-like methods is  to sample from Bayesian posteriors~\citep{deleu2022bayesian,deleu2023joint, atanackovic2023dyngfn}, GFlowNets find varied applications in combinatorial optimization~\citep{Zhang2023}, multi-objective active learning~\citep{mogfn}, and design of biological sequences \citep{sequence}. 
Inheriting jargon from reinforcement learning, GFlowNets learn to sample from $R : \mathcal{X} \rightarrow \mathbb{R}^+$ by incrementally refining a policy function $p_F$ that incrementally builds objects $x \in \mathcal{X}$ by augmenting an initial state $s_0$. In Bayesian context, $R$ is proportional to some posterior $p(\cdot | \mathcal{D})$, and $\mathcal{X}$ is its support.

Nonetheless, similarly to running MCMC chains, training GFlowNets requires repeatedly evaluating $R$. For posterior sampling, this implies repeated sweeps through the data. Furthermore, if the data is scattered across many clients --- e.g., in Bayesian federated learning~\citep{el-mekkaoui_652, vono22} ---, the multiple communication rounds between clients and the server may be a bottleneck.

In the realm of MCMC, \emph{embarrassingly parallel} methods~\citep[][]{Neiswanger2014} address both aforementioned issues simultaneously through a divide-and-conquer scheme. In a nutshell, given an $N$-partition $\mathcal{D}_1, \ldots, \mathcal{D}_N$ of the data $\mathcal{D}$,
the strategy consists of sampling in parallel from $N$ \emph{subposteriors} defined on different data shards:
\begin{equation*}
    R_n(x) \propto p(\mathcal{D}_n | x) p(x)^{1/N} \quad \forall n=1\ldots N,
\end{equation*}
and subsequently combining the results in a server to get approximate samples from the full posterior $p(x|\mathcal{D})$, or equivalently from $R \propto R_1 R_2 \ldots R_N$. In federated settings, the data partition reflects how data arises in client devices, and this class of algorithms incurs a single communication round between the clients and the server. However, these methods are tailored towards continuous random variables, and their combination step relies on approximating (or sampling from) the product of sample-based continuous surrogates of the subposteriors, e.g., kernel density estimators~\citep{Neiswanger2014}, Gaussian processes~\citep{Nemeth2018,EPfailures}, or normalizing flows~\citep{Mesquita2019}. Consequently, they are not well-suited to sample from discrete state spaces.

We propose EP-GFlownet, the first embarrassingly parallel sampling method for discrete distributions. EP-GFlowNets start off by learning $N$ local GFlowNets in parallel to sample proportional to their corresponding reward functions $R_1, \ldots, R_N$ and send the resulting models to the server. Then, the server learns a global GFlowNets by enforcing our newly proposed \emph{aggregating balance} (AB) condition, which ensures the global model correctly samples from the product of (sampling distributions of) the local GFlowNets. Notably, sampling from products of GFlowNets is challenging since no simple composition (e.g., avg./product) of local policies induces the product 
distribution \cite{diffusions}.

Towards deriving the AB, we introduce a novel balance criterion to train conventional/local GFlowNets, the \emph{contrastive balance} (CB) condition. Compared to broadly used training criteria~\citep{Foundations, malkin2022trajectory}, CB often leads to faster convergence, which may be attributed to its minimal parametrization, requiring only forward and backward policies. 
We also show that, in expectation, minimizing the CB is equivalent to minimizing the variance of an estimator 
of the log-partition from \citet{robust}.

Remarkably, the applicability of EP-GFlowNets goes beyond Bayesian inference. For instance, it applies to multi-objective tasks in which we need to sample from a combination of different objective functions \citep{Daulton2021, mogfn}. EP-GFlowNets can also be applied for model reuse~\citep{sculpting}, allowing sampling from log-pools of experts without retraining individual models.

Our experiments validate EP-GFlowNets in different contexts, including multi-objective multiset, parallel Bayesian phylogenetic inference, and federated Bayesian structure learning. In summary, our \textbf{contributions} are: 
\looseness=-1 
\begin{enumerate}[itemsep=0pt,leftmargin=16pt]
    \item We propose EP-GFlowNet, the first algorithm for embarrassingly parallel sampling in discrete state spaces using GFlowNets.
    We provide theoretical guarantees of correctness, and also analyze EP-GFlowNet's robustness to errors in the estimation of local GFlowNets; 
    \item We present the contrastive balance (CB) condition, show it is a sufficient and necessary condition for sampling proportionally to a reward, and analyze its connection to variational inference (VI); 
    \item We substantiate our methodological contributions with experiments on five different tasks. Our empirical results $i$) showcase the accuracy of EP-GFlowNets; $ii$) show that, in some cases, using the CB as training criterion leads to faster convergence compared to popular loss functions; $iii$) illustrate EP-GFlowNets' potential in two notable applications: Bayesian phylogenetic inference, and Bayesian network structure learning. 
\end{enumerate}

\section{Preliminaries} 
\label{sec:gflownets} 

\noindent\textbf{Notation.} We represent a \textit{directed acyclic graph} (DAG) over  nodes $V$ and with adjacency matrix ${A} \in \{1, 0\}^{|V| \times |V|}$ as $G = (V, A)$. 
A \textit{forward policy} over $V$ in $G$ is a function $p \colon V \times V \rightarrow \mathbb{R}_{+}$ such that (i) $p(v, \cdot)$ is a probability measure over $V$ for every $v \in V$ and (ii) $p(v, w) > 0$ if and only if $A_{vw} = 1$; we alternatively write $p(v\rightarrow w)$ and $p(w|v)$ to represent $p(v, w)$. A transition kernel $p$ induces a conditional probability measure over the space of trajectories in $G$: if $\tau = (v_{1} \rightarrow \cdots \rightarrow v_{n})$ is a trajectory of length $n$ in $G$, then $p(\tau | v_{1}) = \prod_{i=1}^{n - 1} p(v_{i + 1} | v_{i})$. A \textit{backward policy} in $G$ is a forward policy on the transpose graph $G^\transpose = (V, A^{\transpose})$.

\noindent\textbf{Generative flow networks.} GFlowNets are a family of amortized variational algorithms trained to sample from an unnormalized distribution over discrete and compositional objects. More specifically, let $R \colon \mathcal{X} \rightarrow \mathbb{R}_{+}$ be an unnormalized distribution over a finite space $\mathcal{X}$. We call $R$ a \textit{reward} due to terminological inheritance from the reinforcement learning literature. Define a finite set $\mathcal{S}$ and a variable $s_{o}$. Then, let $G$ be a weakly connected DAG with nodes $V = \{s_{o}\} \cup \{s_{f}\} \cup \mathcal{S} \cup \mathcal{X}$ such that (i) there are no incoming edges to $s_{o}$, (ii) there are no outgoing edges exiting $s_{f}$ and (iii) there is an edge from each $x \in \mathcal{X}$ to $s_{f}$. We call the elements of $V$ \textit{states} and refer to $\mathcal{X}$ as the set of \textit{terminal states}; $s_{o}$ is called the \textit{initial state} and $s_{f}$ is an absorbing state designating the end of a trajectory. We denote by $\mathcal{T}$ the space of trajectories in $G$ starting at $s_{o}$ and ending at $s_{f}$. Illustratively, $\mathcal{X}$ could be the space of biological sequences of length $32$; $\mathcal{S}$, the space of sequences of lengths up to $32$; and $s_{o}$, an empty sequence. The training objective of a GFlowNet is to learn a forward policy $p_{F}$ over $G$ such that the marginal distribution $p_{T}$ over $\mathcal{X}$ satisfies 
\begin{equation} \label{eq:aaa} 
    p_{T}(x) \coloneqq \sum_{\tau \colon \tau \text{ leads to } x} p_{F}(\tau | s_{o})  \propto R(x).  
\end{equation}
We usually parameterize $p_{F}$ as a neural network and select one among the diversely proposed training criteria to estimate its parameters, which we denote by $\phi_F$. These criteria typically enforce a balance condition on the Markovian process defined by $p_{F}$ that provably imposes the desired property on the marginal distribution in \Cref{eq:aaa}. For example, the \textit{trajectory balance} (TB) criterion introduces a parametrization of the target distribution's partition function $Z_{\phi_Z}$ and of a backward policy $p_{B}(\cdot, \cdot; \phi_B)$ with parameters $\phi_Z$ and $\phi_B$, respectively, and is defined as  
\begin{equation} \label{eq:tbcriterion}  
    \!\!\!\!\!\!
    \begin{split} 
&\mathcal{L}_{TB}(\tau, \phi_F, \phi_B, \phi_Z) = \\& \bigg(\log Z_{\phi_Z} - \log R(x) +   \sum_{s \rightarrow s' \in \tau} \log \frac{p_{F}(s, s'; \phi_F)}{p_{B}(s', s; \phi_B)} \bigg)^{2} 
    \end{split}.     \!\!
\end{equation}

Minimizing \Cref{eq:tbcriterion} enforces the TB condition: $p_{F}(\tau; \phi_F) = Z_{\phi_Z}^{-1}R(x)\prod p_{B}(s', s; \phi_B)$, which implies ~\Cref{eq:aaa} if valid for all $\tau \in \mathcal{T}$. This is the most widely used training scheme for GFlowNets. In practice, some works set $p_B$ as a uniform distribution to avoid learning $\phi_B$, as suggested by \citet{malkin2022trajectory}. 

Another popular approach for training GFlowNets uses the notion of \textit{detailed balance} (DB). Here, we want to find forward and backward policies and \textit{state flows} $F$ (with parameters $\phi_S$) that satisfy the DB condition: $F(s; \phi_S) p_{F}(s, s'; \phi_F) = F(s'; \phi_S) p_{B}(s', s; \phi_B)$ if $s$ is an non-terminal state and $F(s ; \phi_{S}) p_{F}(s_{f} | s; \phi_{F}) = R(s)$ otherwise. 
Again, satisfying the DB condition for all edges in $G$  entails ~\Cref{eq:aaa}.
Naturally, this condition leads to a transition-decomposable loss 
\begin{equation} 
\begin{split}
    \mathcal{L}_{DB}&(s, s', \phi_F, \phi_B, \phi_S) = \\& \!\!\!\!
    \begin{cases} 
    \left(\log \frac{p_{F}(s, s'; \phi_F)}{p_{B}(s', s; \phi_B)} + \log \frac{F(s ; \phi_S)}{F(s' ; \phi_S)} \right)^{2} \text{ if } s' \neq s_{f}, \\
    \left( \log \frac{F(s ; \phi_{S}) p_{F}(s_{f} | s ; \phi_{F})}{R(s)} \right)^{2} \text{otherwise.}  \label{eq:db_criterion} 
    \end{cases}
\end{split}\!\!
\end{equation}

Recently, \citet{LingTrajectory} proposed to residually reparameterize $F$ with reference on a hand-crafted function $\xi$, namely, $\log F(s) = \log \tilde{F}(s) + \log \xi(s)$. This approach led to the \emph{forward-looking (FL) GFlowNet}, whose learning objective we denote by $\mathcal{L}_{FL}$. Importantly, contrarily to \citet{LingTrajectory}, we do not drop the boundary conditions in \autoref{eq:db_criterion}. To guarantee correctness whether using $\mathcal{L}_{DB}$ or $\mathcal{L}_{TB}$, we need to integrate the loss over some exploratory policy $\pi$ fully supported in $\mathcal{T}$. In practice, $\pi$ is typically a $\epsilon$-mixture between $p_{F}$ and a uniform forward policy, $(1 - \epsilon) \cdot p_{F} + \epsilon \cdot u_{F}$, or a tempered version of $p_{F}$. We use the former definition for $\pi$ in this work. We review alternative training schemes for GFlowNets in the supplementary material.
\looseness=-1




\noindent\textbf{Problem statement.} Given a set of clients $n=1,\ldots,N$, each with reward function $R_n$, we want to learn a GFlowNet to sample proportionally to a global reward function $R$ defined as a product of the local rewards $R_1, \ldots, R_N$. We want to do so with as little client-server communication as possible and without openly disclosing the local rewards to the server.
While we focus on sampling from $R(x) \defeq \prod_{n=1}^{N} R_n(x)$ in the main paper, we also provide in \Cref{sec:app:exp} extensions of our theoretical results to exponentially weighted rewards of the form $R(x) \defeq \prod_{n=1}^{N} R_n(x)^{w_n}$ with $w_1, \ldots, w_N > 0$.

\section{Method} 
\label{sec:method} 

This section derives a provably correct framework for embarrassingly parallel GFlowNets based on the newly developed concept of \emph{aggregating balance} (\Cref{subsec:federated}).  
As a cornerstone, we introduce the \emph{contrastive balance (CB) condition}, a new balance condition requiring minimal parametrization. Notably, \Cref{subsec:contrastive} shows the CB condition yields a sound learning objective for training conventional GFlowNets.


\subsection{Embarrassingly Parallel GFlowNets}
\label{subsec:federated}


To circumvent the restrictions imposed by the problem statement, we propose a divide-and-conquer scheme. First, each client trains their own GFlowNet to sample proportionally to their local reward, sending the estimated forward and backward policies $p_F^{(n)}$ and $p_B^{(n)}$ to a centralizing server. Then, the server estimates the policies $(p_F, p_B)$ of a novel GFlowNet that approximately samples from $R$ solely based on the local policies $\{(p_{F}^{(n)}, p_{B}^{(n)})\}_{n=1}^{N}$, i.e., without ever evaluating any $R^{(n)}$. More specifically, the server learns a GFlowNet whose marginal distribution $p_T$ over terminal states is proportional to the product of those from the clients $p_T^{(1)}, \ldots, p_T^{(N)}$. 
Toward this end, \Cref{thm:federated_condition} delineates a necessary and sufficient condition guaranteeing the correctness of the aggregation phase, which we call \emph{aggregating balance (AB) condition}. It is worth mentioning that \Cref{thm:federated_condition} builds directly upon the contrastive balance condition in \cref{lemma:contrastive}, discussed in detail in \Cref{subsec:contrastive}. 

\begin{theorem}[Aggregating balance condition]
\label{thm:federated_condition}    Let $\left(p_F^{(1)}, p_{B}^{(1)}\right),  \dots, \left(p_F^{(N)}, p_{B}^{(N)}\right): V^2 \rightarrow \mathbb{R}^+$ be pairs of forward and backward policies from $N$ GFlowNets sampling respectively proportionally to $R_1, \ldots, R_N  : \mathcal{X} \rightarrow \mathbb{R}^+$.
    Then, another GFlowNet with forward and backward policies $p_F, p_B \in V^2 \rightarrow \mathbb{R}^+$ samples proportionally to $R(x) \defeq \prod_{n=1}^{N} R(x)$ if and only if the following condition holds for all terminal  trajectories $\tau, \tau^\prime \in \mathcal{T}$:
    \begin{equation}
        \!\!\prod_{n=1}^{N} \frac{\left(\displaystyle\prod_{s \rightarrow s' \in \tau} \frac{p_{F}^{(i)}(s, s')}{p_{B}^{(i)}(s', s)} \right)} {\left(\displaystyle\prod_{s \rightarrow s' \in \tau'} \frac{p_{F}^{(i)}(s, s')}{p_{B}^{(i)}(s', s)}\right)} = 
        \frac{\left(\displaystyle\prod_{s \rightarrow s' \in \tau} \frac{p_{F}(s, s')}{p_{B}(s', s)} \right)}{\left(\displaystyle\prod_{s \rightarrow s' \in \tau'} \frac{p_{F}(s, s')}{p_{B}(s', s)}\right)}. 
        \!\!
        \label{eq:fed_balance}
    \end{equation}
\end{theorem}

Based on \Cref{thm:federated_condition}, we can naturally derive a loss function enforcing \Cref{eq:fed_balance} that can be used to combine the locally trained GFlowNets. To guarantee the minimum of our loss achieves aggregating balance, it suffices to integrate \Cref{eq:fed_balance} against a distribution attributing non-zero mass to every $(\tau, \tau') \in \mathcal{T}^2$. Importantly, note the aggregating balance loss is agnostic to which loss was used to learn the local GFlowNets, as it only requires their transition functions and not, e.g., an estimate of the partition function or of the flows going though each state.

\begin{figure*}[t]
    \centering
    \includegraphics[width=0.9\textwidth]{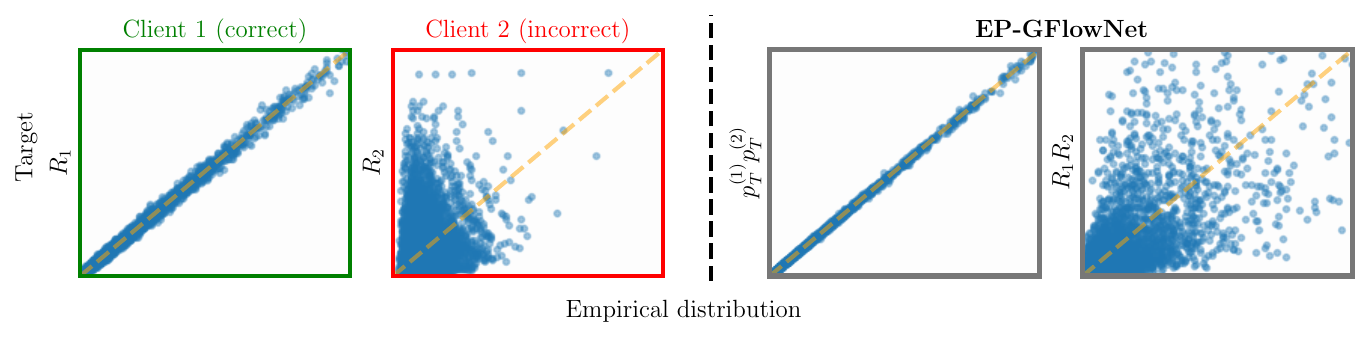}
    \caption{\textbf{EP-GFlowNet samples proportionally to a pool of locally trained GFlowNets.} If a client correctly trains their local model (\textcolor{teal}{green}) and another client trains theirs incorrectly (\textcolor{red}{red}), the distribution inferred by EP-GFlowNet (mid-right) differs from the target product distribution (right)\looseness=-1.}
    \label{fig:illustration}
\end{figure*}

\begin{corollary}[Aggregating balance loss] Let $p^{(i)}_{F}$ and $p^{(i)}_{B}$ be forward/backward transition functions such that $p^{(i)}_T(x) \propto R_i(x)$ for arbitrary reward functions $R_i$ over terminal states $x \in \mathcal{X}$. Also, let $\nu : \mathcal{T}^2 \rightarrow \mathbb{R}^+$ be a full-support distribution over pairs of complete trajectories. Moreover, assume that $p_F(\cdot, \cdot ; \phi_F)$ and $p_B(\cdot, \cdot ; \phi_B)$ denote the forward/backward policies of a GFlowNet parameterized by $\phi_F$ and $\phi_B$. The following are equivalent:
\begin{enumerate}[leftmargin=16pt]
\item $p_{T}(x; \phi_F) \propto \prod_i R_i(x)$ for all $x \in \mathcal{X}$;
\item $\mathbb{E}_{(\tau, \tau') \sim \nu} \left[ \mathcal{L}_{AB}(\tau, \tau', \phi_F, \phi_B) \right]=0$ where for trajectories $\tau \rightsquigarrow x$ and $\tau \rightsquigarrow x'$
\end{enumerate}
\begin{equation} \label{eq:aaaaaaaa} 
    \begin{split} 
    \mathcal{L}_{AB}(\tau, \tau', \phi_{F}, \phi_{B}) \!\! = \!\! \bigg( \log \frac{p_{F}(\tau ; \phi_{F}) p_{B}(\tau' | x'; \phi_{B}) }{p_{B}(\tau | x ; \phi_{B}) p_{F}(\tau' ; \phi_{F})} \\ - \!\!\!\! \sum_{1 \le i \le N} \!\!\! \log \frac{p_{F}(\tau ; \phi_{F}) p_{B}(\tau' | x'; \phi_{B}) }{p_{B}(\tau | x ; \phi_{B}) p_{F}(\tau' ; \phi_{F})} \bigg)^{2}.
    \end{split} 
\end{equation}
\end{corollary}



\begin{remark}[Imperfect local inference] \label{remark1}
    In practice, the local balance conditions often cannot be completely fulfilled by the local GFlowNets and the distributions $p_T^{(1)}, \ldots, p_T^{(N)}$  over terminal states 
    are not proportional to the rewards $R_{1}, \ldots, R_N$. In this case,  aggregating balance implies the aggregated model  samples proportionally to 
    \begin{equation}         \mathbb{E}_{\tau \sim p_{B}(\cdot | x)} \left[ \prod_{1 \le i \le N} \frac{p^{(i)}_{F}(\tau)}{p^{(i)}_{B}(\tau | x)} \right]. \label{eq:expected_prod_reward}
    \end{equation}

    Interestingly, the value of \Cref{eq:expected_prod_reward} equals the expectation of a non-deterministic random variable only if the local balance conditions are not satisfied. Otherwise, the ratio $\nicefrac{p^{(i)}_{F}(\tau)}{p^{(i)}_{B}(\tau | x)}$ equals $R(x)$, a constant wrt $\tau$ conditioned on $\tau$ having $x$ as its final state. Furthermore, \Cref{eq:expected_prod_reward} also allows us to assess the probability mass function the global EP-GFlowNet is truly drawing samples from. 
\end{remark}

As mentioned in \cref{remark1}, in practice, the local GFlowNets may not be balanced with respect to their rewards, incurring errors that propagate to our aggregated model. In this context, \Cref{thm:robustness} quantifies the extent to which these local errors impact the overall result. 

\begin{theorem}[Influence of local failures] \label{thm:robustness} 
   Let $\pi_n \defeq R_{n}/Z_{n}$ and $p_{F}^{(n)}$ and $p_{B}^{(n)}$ be the forward and backward policies of the $n$-th client. We use $\tau \rightsquigarrow x$ to indicate that $\tau \in \mathcal{T}$ is finished by $x \rightarrow s_{f}$. Suppose that the local balance conditions are lower- and upper-bounded $\forall\, n \in [[1, N]]$ as  
   \begin{equation}
        \begin{split} 
           1 - \alpha_{n} \le \min_{x \in \mathcal{X}, \tau \rightsquigarrow x} \frac{p_{F}^{(n)}(\tau)}{p_{B}^{(n)}(\tau | x) \pi_{n}(x)} \\ \le \max_{x \in \mathcal{X}, \tau \rightsquigarrow x} \frac{p_{F}^{(n)}(\tau)}{p_{B}^{(n)}(\tau | x) \pi_{n}(x)} \le 1 + \beta_{n} 
       \end{split} 
   \end{equation}
    where $\alpha_{n} \in (0, 1)$ and $\beta_{n} > 0$. The Jeffrey divergence $\mathcal{D}_{J}$ between the global model $\hat{\pi}(x)$ that fulfills the aggregating balance condition and $\pi(x) \propto \prod_{n=1}^{N} \pi_{n}(x)$ then satisfies 
    \begin{equation}
        \mathcal{D}_{J}(\pi, \hat{\pi}) \le \sum_{n=1}^{N} \log\left(\frac{1 + \beta_{n}}{1 - \alpha_{n}}\right). \label{eq:jeff_bound} 
    \end{equation}
\end{theorem}

There are two things worth highlighting in \Cref{thm:robustness}. First, if the local models are accurately learned (i.e., $\beta_n = \alpha_n = 0\, \forall n$), the RHS of \Cref{eq:jeff_bound} equals zero, implying $\pi = \hat{\pi}$.
Second, if either $\beta_n \rightarrow \infty$ or $\alpha_n \rightarrow 1$ for some $n$, the bound in \Cref{eq:jeff_bound} goes to infinity --- i.e., it degenerates if one of the local GFlowNets are poorly trained. This is well-aligned with the \textit{catastrophic failure} phenomenon~\citep{EPfailures}, which was originally observed in the literature of parallel MCMC~\citep{Neiswanger2014, Nemeth2018, Mesquita2019} and refers to the incorrectness of the global model due to inadequately estimated local parameters and can result in missing modes or misrepresentation of low-density regions. 
\Cref{fig:illustration} shows a case where one of the local GFlowNets is poorly trained (Client 2's). Note that minimizing the AB objective leads to a good approximation of the product of marginal distributions over terminal states (encoded by the local GFlowNets). Nonetheless, the result is far from we have envisioned at first, i.e., the learned model significantly diverges from the product distribution $R \propto R_1 R_2$. 
Additionally, \Cref{fig:app:clients} in \Cref{sec:app:moreexp} highlights that EP-GFlowNets can learn a relatively good approximation to the target distribution even in the face of inaccurate local approximations. 
\looseness=-1 

\subsection{Contrastive balance} 
\label{subsec:contrastive}

As a stepping stone towards proving \Cref{thm:federated_condition}, we develop the \emph{contrastive balance condition}, which is sufficient for ensuring that a GFlowNet's marginal over terminal states is proportional to its target reward (\cref{lemma:contrastive}). 

\begin{lemma}[Constrastive balance condition]
\label{lemma:contrastive}
If $p_F, p_B \in V^2 \rightarrow \mathbb{R}^{+}$ are the forward and backward policies of a GFlowNet sampling proportionally to some arbitrary reward function $R : \mathcal{X} \rightarrow \mathbb{R}^+$, then, for any pair of complete trajectories $\tau, \tau^\prime \in \mathcal{T}$ with $\tau \rightsquigarrow x$ and $\tau \rightsquigarrow x'$, 
\begin{equation}
    {R(x')} \prod_{s \rightarrow s' \in \tau} \frac{p_{F}(s, s')}{p_{B}(s', s)} 
       = {R(x)} \prod_{s \rightarrow s' \in \tau'} \frac{p_{F}(s, s')}{p_{B}(s', s)}. \label{eq:contrastive_balance_condition}
\end{equation}
Conversely, if a GFlowNet with forward and backward policies $p_F, p_B$ abide by \Cref{eq:contrastive_balance_condition}, it induces a marginal distribution over $x 
\in \mathcal{X}$ proportional to $R$. 
\end{lemma}


Enforcing \cref{lemma:contrastive} results in a loss that does not depend on an estimate $\log Z_{\phi_Z}$ of the intractable log-partition function present in the TB condition. The next corollary guarantees that an instantiation of the GFlowNet parameterized by a global minimizer of $\mathcal{L}_{CB}$ (\Cref{eq:contrastive_loss}) correctly samples from  $p(x) \propto R(x)$. We call $\mathcal{L}_{CB}$ the contrastive balance loss as it measures the contrast between randomly sampled trajectories. 
In practice, we observed that in some cases the CB loss leads to better results than the TB and DB losses, as we will see in   \Cref{subsec:cb_exp}.

\begin{corollary}[Contrastive balance loss] \label{col:contrastive} Let $p_F(\cdot, \cdot ; \phi_F)$ and $p_B(\cdot, \cdot ; \phi_B)$ denote forward/backward policies,
and $\nu : \mathcal{T}^2 \rightarrow \mathbb{R}^+$ be a full-support probability distribution over pairs of terminal trajectories. Then, $p_{T}(x; \phi_F) \propto R(x)$ $\forall x \in \mathcal{X}$ iff $\mathbb{E}_{(\tau, \tau') \sim \nu} \left[ \mathcal{L}_{CB}(\tau, \tau', \phi_F, \phi_B) \right]=0$ where
\begin{equation} \label{eq:contrastive_loss} 
    \begin{split}
        \mathcal{L}_{CB}(\tau, \tau',  \phi_{F}, \phi_{B}) = \bigg( \log \frac{p_{F}(\tau ; \phi_{F})}{p_{B}(\tau ; \phi_{B})} - \\ \log \frac{p_{F}(\tau' ; \phi_{F})}{p_{B}(\tau' ; \phi_{B})} + \log \frac{R(x')}{R(x)} \bigg)^{2}.  
    \end{split}
\end{equation} 
\end{corollary}

 \noindent\textbf{Computational advantages of $\mathcal{L}_{CB}$.}  Importantly, note that $\mathcal{L}_{CB}$ incurs learning fewer parameters than TB and DB losses. Indeed, besides requiring the forward and backward policies $p_{F}$ and $p_{B}$, TB requires parameterizing the partition function of $R$. Alternatively, DB implies using a neural network to approximate the flow through each node. In contrast, CB requires only learning $p_F$ and $p_B$. 

\noindent\textbf{$\mathcal{L}_{CB}$ and VI.} Notably, the next proposition ties the CB loss' gradient to that of a variational objective, extending the  characterization of GFlowNets as VI started by \citet{malkin2023gflownets} for the TB loss. More specifically, \Cref{prop:variational_objective} states that the on-policy gradients of the CB objective coincide in expectation to the gradient of the KL divergence between the forward and backward policies. 

\begin{theorem}[VI \& CB] \label{prop:variational_objective} 
    Let $p_{F} \otimes p_{F}$ be the outer product distribution assigning probability $p_{F}(\tau)p_{F}(\tau')$ to each trajectory pair $(\tau, \tau')$. The criterion in \Cref{eq:contrastive_loss} satisfies 
    \begin{equation*}
        \nabla_{\phi_F} \mathcal{D}_{KL}[p_{F} || p_{B}] = \frac{1}{4} \underset{{(\tau, \tau') \sim p_{F} \otimes p_{F}}}{\mathbb{E}} \!\!\!\!\! \left[ \nabla_{\phi_F} \mathcal{L}_{CB}(\tau, \tau', \phi_F) \right].  
    \end{equation*}
\end{theorem}

Notably, a corresponding result connecting the CB's and KL's gradients holds when we parameterize the backward policy, as we show in the Theorem [ref] in the appendix.  

\noindent\textbf{Connection to other balance conditions.} The CB loss may be equivalently defined as the squared difference between signed violations to the TB of two independent trajectories, namely, $\mathcal{L}_{CB}(\tau, \tau') = \left( \mathcal{V}_{TB}(\tau) - \mathcal{V}_{TB}(\tau') \right)^{2}$, with $\mathcal{V}_{TB}(\tau) = \log \left(\nicefrac{p_{F}(\tau) Z}{p_{B}(\tau | x) R(x)}\right)$ satisfying $\mathcal{V}_{TB}(\tau)^{2} = \mathcal{L}_{TB}(\tau)$. In this scenario, one might derive \cref{lemma:contrastive} and \cref{prop:variational_objective} as corollaries of the respective results for the TB condition by \citet[Proposition~1]{malkin2022trajectory} and by \citet[Proposition~1]{malkin2023gflownets}. We present this derivation --- jointly with self-contained proofs --- in \Cref{sec:app:p} and \Cref{sec:app:cb}. 
\Cref{sec:app:cb} 
also shows that the expected value of the CB loss is proportional to the variance of a TB-based estimate of the partition function, which was used as a training loss by \citet{robust}. 


\begin{table*}[h!] 
    \caption{\textbf{Quality of the parallel approximation} to the combined rewards. The table shows i) the L1 distance between the distribution induced by each method and the ground truth and ii) the average log reward of the top-800 scoring samples. Our EP-GFlowNet is consistently better than the PCVI baseline regarding L1 distance, showing approximately the same performance as a centralized GFlowNet. Furthermore, EP-GFlowNet's Top-800 score perfectly matches the centralized model, while PCVI's differ drastically. Values are the average and standard deviation over three repetitions.}
    \centering
    \begin{adjustbox}{center, max width=0.9\textwidth}
    \centering
    \begin{tabular}{c c cc cc c}
             \toprule 
         & \multicolumn{2}{c}{\textbf{Grid World}} & \multicolumn{2}{c}{\textbf{Multisets}} & \multicolumn{2}{c}{\textbf{Sequences}} \\ 
         & $L_{1} \downarrow$ & Top-800 $\uparrow$ & $L_{1} \downarrow$ & Top-800 $\uparrow$ & $L_{1} \downarrow$ & Top-800 $\uparrow$ \\ 
         \midrule 
         \multirow{2}{*}{Centralized} & 

          $0.027$ 
          & $-6.355$ 
          & $0.100$ 
          & $27.422$ 
          & $0.003$ 
          & $-1.535$ \\ 
         & $\textcolor{gray}{ {\scriptstyle( \pm 0.016) } }$
         & $\textcolor{gray}{ {\scriptstyle( \pm 0.000) } }$
         & $\textcolor{gray}{ {\scriptstyle( \pm 0.001) } }$
         & $\textcolor{gray}{ {\scriptstyle( \pm 0.000) } }$
         & $\textcolor{gray}{ {\scriptstyle( \pm 0.001) } }$
         & $\textcolor{gray}{ {\scriptstyle( \pm 0.000) } }$ \\[0.1cm] 
        \hdashline \noalign{\vskip 0.1cm} 
         \multirow{2}{*}{EP-GFlowNet (\textbf{ours})} & 

        $\mathbf{0.038}$ 
         & $\mathbf{-6.355}$ 

         & $\mathbf{0.130}$
         & $\mathbf{27.422}$ 

         & $\mathbf{0.005}$
         & $\mathbf{-1.535}$ \\ 

         & $\textcolor{gray}{ {\scriptstyle( \pm 0.016) } }$
         & $\textcolor{gray}{ {\scriptstyle( \pm 0.000) } }$

         & $\textcolor{gray}{ {\scriptstyle( \pm 0.004) } }$
         & $\textcolor{gray}{ {\scriptstyle( \pm 0.000) } }$ 

         & $\textcolor{gray}{ {\scriptstyle( \pm 0.002) } }$
         & $\textcolor{gray}{ {\scriptstyle( \pm 0.000) } }$ \\ 
        
         \multirow{2}{*}{PCVI} & 

         $0.189$ 
         & $\mathbf{-6.355}$

         & $0.834$ 
         & $26.804$ 

         & $1.872$
         & $-16.473$ \\ 

         & $\textcolor{gray}{ {\scriptstyle( \pm 0.006) } }$
         & $\textcolor{gray}{ {\scriptstyle( \pm 0.000) } }$

         & $\textcolor{gray}{ {\scriptstyle( \pm 0.005) } }$
         & $\textcolor{gray}{ {\scriptstyle( \pm 0.018) } }$

         & $\textcolor{gray}{ {\scriptstyle( \pm 0.011) } }$ 
         & $\textcolor{gray}{ {\scriptstyle( \pm 0.007) } }$ \\
        \bottomrule 
    \end{tabular}
    \end{adjustbox} 
    \vspace{-8pt} 
    \label{tab:gflownets}
\end{table*}

\begin{figure*}[t!] 
    \centering 
    \includegraphics[width=0.75\textwidth]{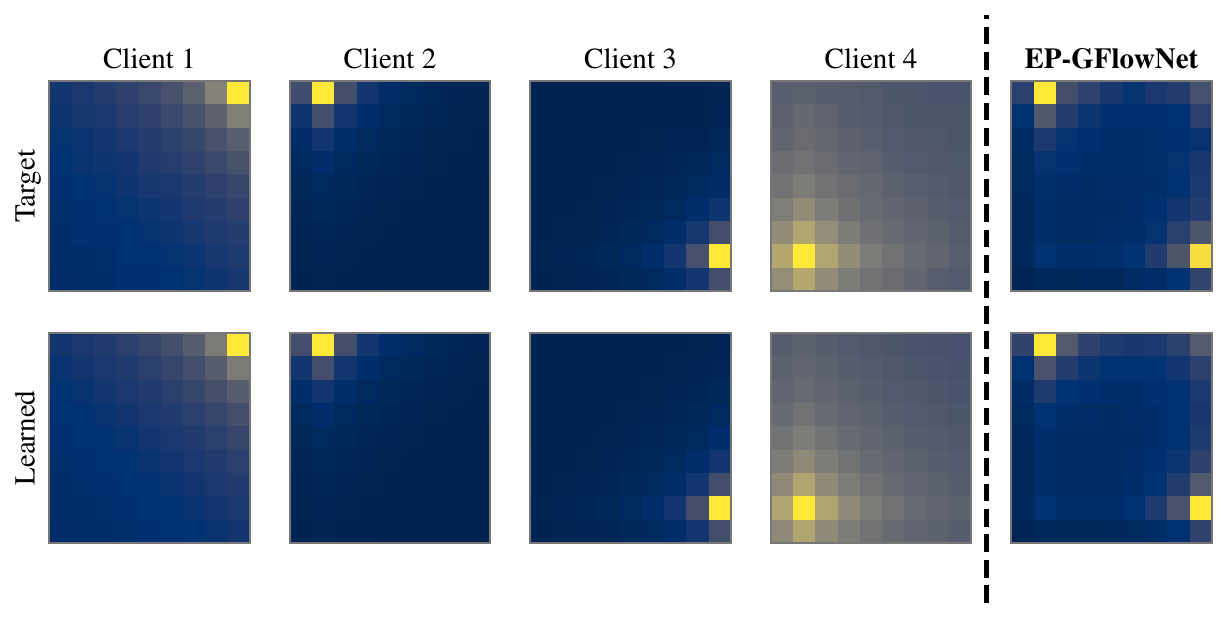} 
    \vspace{-14pt} 
    \caption{\textbf{Grid world.} Each heatmap represents the target distribution (first row), based on the normalized reward, and the ones learned by the local GFlowNets (second row). Results for EP-GFlowNet are in the rightmost panels. As established by \Cref{thm:federated_condition}, the good fit of the local models results in an accurate fit to the combined reward.}
    \vspace{-8pt}
\label{fig:grid}
\end{figure*}

\section{Experiments} \label{sec:experiments} 

The main purpose of our experiments is to verify the empirical performance of EP-GFlowNets, i.e., their capacity to accurately sample from the combination of local distributions. To that extent, we consider five diverse tasks: sampling states from a \textit{grid world} in \Cref{subsec:grid}, \textit{generation of multisets}~\citep{Foundations, LingTrajectory} in Section \ref{subsec:multiset}, \textit{design of sequences} \citep{sequence} in \Cref{subsec:sequence}, \textit{distributed Bayesian phylogenetic inference}~\citep{matsen} in \Cref{subsec:phylo}, and \textit{federated Bayesian network structure learning} \citep[BNSL;][]{Ng2022federated} in \Cref{sec:dags}. Since EP-GFlowNet is the first of its kind, we propose two baselines to compare it against: a centralized GFlowNet, which requires clients to disclose their rewards in a centralizing server, and a divide-and-conquer algorithm in which each client approximates its local GFlowNet with a product of categorical distributions, which are then aggregated in the server. We call the latter approach parallel categorical VI (PCVI), which may be also viewed as an implementation of the SDA-Bayes framework for distributed approximate Bayesian inference \cite{tamara}. 

\subsection{Grid world} 
\label{subsec:grid}

\noindent\textbf{Task description.} Our grid world environment consists of a Markov decision process over an  $9\times9$ square grid in which actions consist of choosing a direction ($\rightarrow, \uparrow$) or stopping. The reward for each state $R$ is the sigmoid transform of its minimum distance to a reward beacon (bright yellow in \Cref{fig:grid}). For the distributed setting, we consider the problem of combining the rewards from different clients, each of which has two beacons placed in different positions.

\noindent\textbf{Results.} Figure~\ref{fig:grid} shows that EP-GFlowNet accurately approximates the targeted distribution, even in cases where combining the client rewards leads to multiple sparsely distributed modes. Furthermore, \Cref{tab:gflownets} shows that EP-GFlowNet performs approximately on par with the centralized model --- trained on the product distribution --- in terms of $L_1$ distance (within one standard deviation), but is three orders of magnitude better than the PCVI baseline. This is also reflected in the average reward over the top 800 samples --- identical to the centralized version for EP-GFlowNet, but an order of magnitude smaller for PCVI. Again, these results corroborate our theoretical claims about the correctness of our scheme for combining GFlowNets. 

\begin{figure*}[h] 
    \centering 
    \includegraphics[width=.9\textwidth]{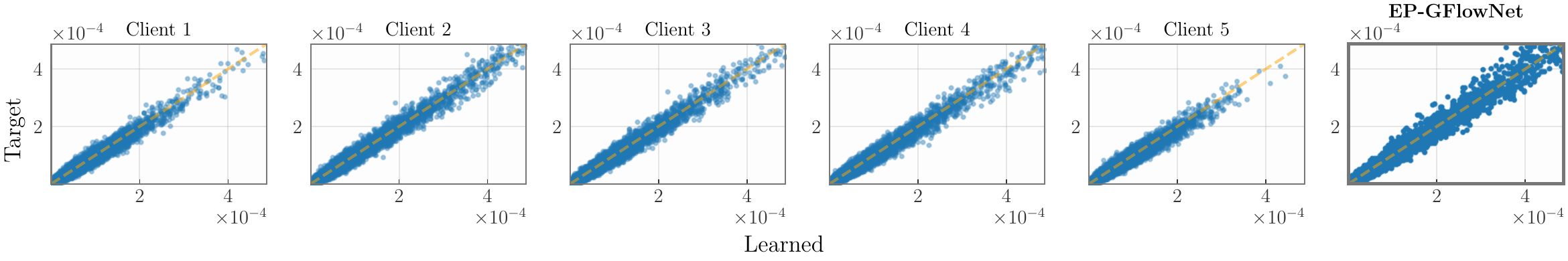} 
    \vspace{-12pt} 
    \caption{\textbf{Multisets: learned $\times$ ground truth distributions.} Plots compare target vs. distributions learned by GFlowNets. The five plots to the left show local models were accurately trained. Thus, a well-trained EP-GFlowNet (right) approximates well the combined reward.} 
    \label{fig:multiscale}
    \vspace{-8pt}
\end{figure*}

\begin{figure*}[t!] 
    \centering 
    \includegraphics[width=.9\textwidth]{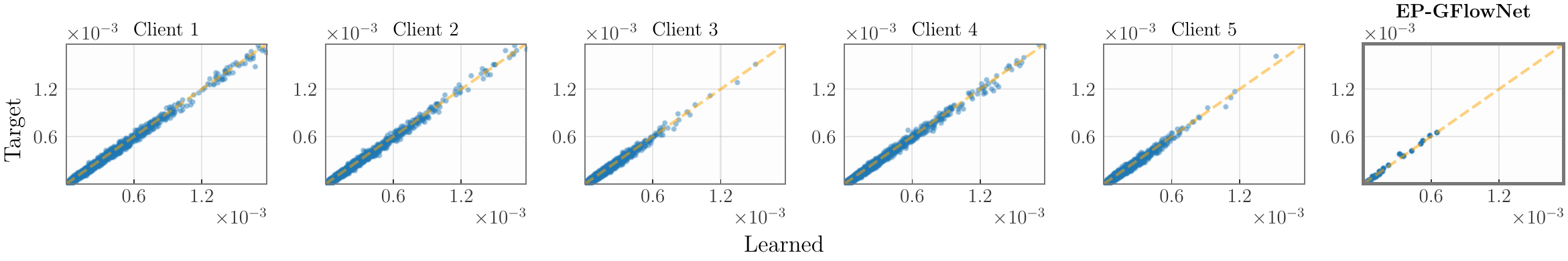} 
    \vspace{-12pt} 
    \caption{\textbf{Sequences: learned $\times$ ground truth distributions.} . Plots compare target to distributions learned. The five leftmost plots show local GFlowNets were well trained. Hence, as implied by \Cref{thm:federated_condition}, EP-GFlowNet approximates well the combined reward.} 
    \label{fig:sequence}
    \vspace{-8pt}
\end{figure*}

\subsection{Multiset generation} 
\label{subsec:multiset}

\noindent\textbf{Task description.} Here, the set of terminal states comprises multisets of size $S$. A multiset $\mathcal{S}$ is built by iteratively adding elements from a finite dictionary $U$ to an initially empty set. Each client $n$ assigns a value $r^{n}_{u}$ to each $u \in U$ and defines the log-reward of $\mathcal{S}$ as the sum of its elements' values; i.e., $\log R_n(\mathcal{S}) = \sum_{u \in \mathcal{S}} r^{(n)}_{u}$. 
In practice, the quantities $r^{(n)}_{u}$ are uniformly picked from the interval $[0, 1]$ for each client. We use $S=8$ and $|U|=10$ in our experiments.

\noindent\textbf{Results.} Figure~\ref{fig:multiscale} provides further evidence that our algorithm is able to approximate well the combined reward, even if only the local GFlowNets are given. This is further supported by the results in \Cref{tab:gflownets}. Notably, EP-GFlowNet is roughly eight times more accurate than the PCVI baseline.


\subsection{Design of sequences} 
\label{subsec:sequence}

\noindent\textbf{Task description.} This tasks revolves around building sequences of maximum size $S$. We start with an empty sequence $\mathcal{S}$ and proceed by iteratively appending an element from a fixed dictionary $U$. The process halts when (i) we select a special terminating token or (ii) the sequence length reaches $S$.  In the distributed setting, we assume each client $n$ has a score $p^{(n)}_{s}$ to each of the $S$ positions within the sequence and a score $t^{(n)}_{u}$ to each of the $|U|$ available tokens,  yielding the logarithmic reward of a sequence $\mathcal{S} = (u_{1}, \dots, u_{M})$ as $\log R_n(\mathcal{S}) = \sum_{i=1}^{M} p^{(n)}_{i} t^{(n)}_{u_{i}}$. 

\noindent\textbf{Results.} Again, Figure~\ref{fig:sequence} corroborates \Cref{thm:federated_condition} and shows that EP-GFlowNet accurately samples from the product of rewards. \Cref{tab:gflownets} further reinforces this conclusion, showing a small gap in $L_{1}$ distance between EP-GFlowNet and the centralized GFlowNet trained with access to all rewards. Notably, our method is $\approx 8\times$ more accurate than PCVI. Furthermore, the Top-800 average reward of EP-GFlowNet perfectly matches the centralized model.

 \subsection{Bayesian phylogenetic inference} 
\label{subsec:phylo}

\noindent\textbf{Task description.} In this task, we are interested on inferring a \textit{phylogeny} $\mathrm{T} = (t, b)$, which is a characterization of the evolutionary relationships between biological species and is composed by a tree topology $t$ and its $(2N-1)$-dimensional vector of non-negative branch lengths $b$. The topology $t$ is as a leaf-labeled complete binary tree with $N$ leaves, each corresponding to a species. Notably, $\mathrm{T}$ induces a probability distribution $P$ over the space of nucleotide sequences $Y_{1}, \dots, Y_{M} \in \Omega^{N}$, where $\Omega$ is a vocabulary of \emph{nucleobases} and $Y_m$ denotes the nucleobases observed at the $m$-th site for each species. Assume $t$ is rooted in some node $r$ and that ${\pi} \in \Delta^{|\Omega|}$ is the prior probability distribution over the nucleobases' frequencies at $r$. Then, the marginal likelihood of a nucleobase $Y_{m}$ occurring \textit{site} ${m}$ for node $n$ is recursively defined by \citet{Felsenstein1981}'s algorithm as 
\begin{equation*}
    \mathbf{P}_{n}(Y_{m} | \mathrm{T}) = 
    \begin{cases}
        \text{One-Hot}(Y_{m,n}) & \text{ if } n \text{ is a leaf,} \\ 
        \left[ \mathbf{M}(n, n_{l}) \odot \mathbf{M}(n, n_{r}) \right]^\intercal & \text{ otherwise}, 
    \end{cases}
\end{equation*}
in which $n_{l}$ and $n_{r}$ are respectively the left and right children of $n$; ${b}_{n, a}$ is the length of the branch between nodes $n$ and $a$; ${Q} \in \mathbb{R}^{|\Omega| \times |\Omega|}$ is an instantaneous rate-conversion matrix for the underlying substitution rates between nucleotides, which is given beforehand; and $\mathbf{M}(n, k) = e^{{b}_{n, k} {Q}} \mathbf{P}_{k}(Y_{m} | \mathrm{T})^\intercal$ represents the mutation probabilities from entity $n$ to entity $k$ after a time $b_{n, k}$. In this context, the marginal likelihood of the observed data within the site m is $\mathbf{P}_{r}(Y_{m} | \mathrm{T})^{\intercal}{\pi}$ and, assuming conditional independence of the sites given $T$, the overall likelihood of the data is $\mathbf{P}(\mathbf{Y} | \mathrm{T}) = \prod_{1 \le i \le M} (\mathbf{P}_{r}(Y_{i} | \mathrm{T})^{\intercal} {\pi})$ --- which is naturally log-additive on the sites. 
For our experiments,  ${\pi}$ is a uniform distribution. For simplicity, we consider constant branch length, fixed throughout the experiments. For the distributed setting, we place a uniform prior over $t$ and split $2500$ nucleotide sites across five clients. In parallel MCMC~\citep{Neiswanger2014} fashion, each client trains a GFlowNet to sample from its local posterior, proportional to the product of its local likelihood and a scaled version of the prior. We further detail the generative process for building phylogenetic trees in \Cref{fig:trees} in \Cref{sec:app:experiments}.  

\begin{figure*}[!t] 
    \centering 
    \includegraphics[width=.9\textwidth]{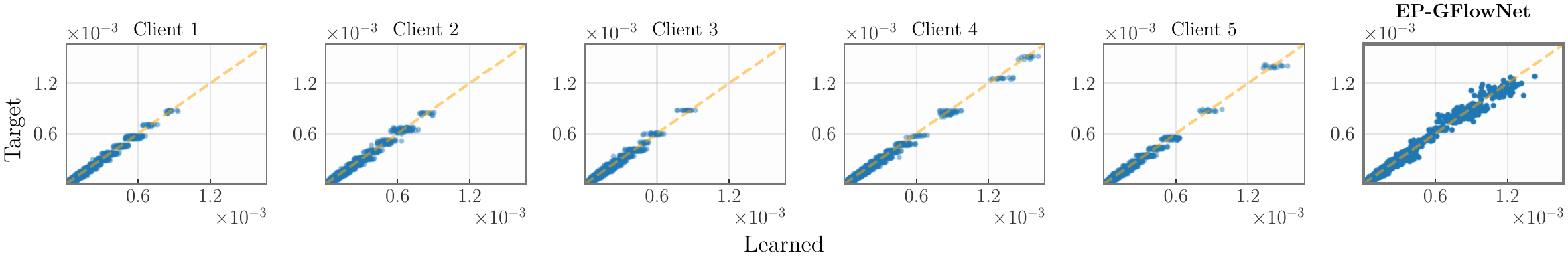} 
    \vspace{-12pt} 
    \caption{\textbf{Bayesian phylogenetic inference: learned $\times$ ground truth distributions.} Following the pattern in Figures \ref{fig:grid}-\ref{fig:sequence}, the goodness-of-fit from local GFlowNets (Clients 1-5) is directly reflected in the distribution learned by EP-GFlowNet.} 
    \label{fig:phylo}
\end{figure*}

\noindent\textbf{Results.} \Cref{fig:phylo} shows that EP-GFlowNet accurately learns the posterior distribution over the tree's topologies: the $L_{1}$ error between the learned distribution and the targeted product distribution is $0.088$, whereas the average $L_{1}$ error among the clients is $0.083 {\scriptstyle( \pm 0.041)}$. Noticeably, this indicates the model's aptitude to learn a posterior distribution in a decentralized manner. Moreover, our results suggest the potential usefulness of GFlowNets as a scalable alternative to the notoriously inefficient MCMC-based algorithms~\citep{matsen} in the field of evolutionary biology. Indeed, \Cref{fig:federated} at \Cref{sec:app:experiments} highlights that our distributed framework significantly reduces the training runtime for GFlowNets for Bayesian inference relatively to a centralized approach.
Notably, naive strategies, like the PCVI baseline, consistently lead to sampling elements that do not belong to the support of our posterior (i.e., are invalid) --- which is why we do not compare against it. 
In future endeavors, we plan to investigate joint parallel inference on the tree's topology and its branches' lengths using hybrid-space GFlowNets ~\citep{deleu2023joint}. Importantly, our method is also the first provably correct algorithm enabling distributed Bayesian inference over discrete objects, which may become invaluable in real-world problems with several thousands of sites \cite{stamatakis2013novel}. 

\begin{figure}
    \centering
    \includegraphics[width=.8\linewidth]{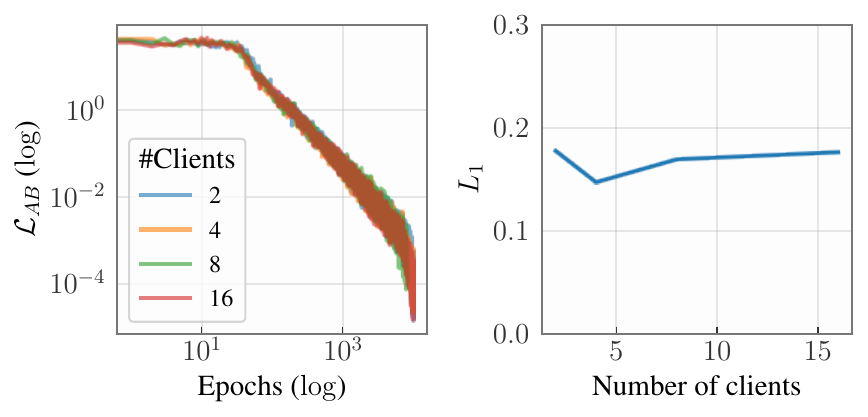}
    \caption{\textbf{EP-GFlowNet are relatively stable} wrt the number of clients. The learning curve of the aggregation phase (left) and the accuracy achieved by the resulting model in terms of $L_{1}$ norm (right) are roughly the same for varying number of clients.}
    \label{fig:increasing}
    \vspace{-12pt} 
\end{figure}

\noindent\textbf{EP-GFlowNets' scalability wrt number of clients.} To illustrate the effect of the number of clients, we perform the training of EP-GFlowNets with an increasing number of clients and a fixed data set. Strikingly, \Cref{fig:increasing} shows that EP-GFlowNets are approximately robust to the number of chunks the data is partitioned into --- both in terms of convergence speed and accuracy.  

\subsection{Federated Bayesian network structure learning} 
\label{sec:dags} 


\noindent\textbf{Task description.} 
While structure learning is usually carried out on centralized data, there are situations in which the possibly sensitive data is scattered across a number of clients~\cite{reisach2021beware, lorch2021dibs}. In this case, local datasets may be small and individually provide insufficient information to draw meaningful conclusions. Thus, we extend the work of  \citet{deleu2022bayesian} and show that EP-GFlowNets can efficiently learn beliefs over Bayesian networks in a federated setting \cite{Ng2022federated}, drawing strength from multiple data sources.
To this end, let $\mathbf{X} \in \mathbb{R}^{d}$ be a random variable and $\mathbf{B} \in \mathbb{R}^{d \times d}$ be a real-valued matrix with sparsity pattern determined by a DAG $\mathcal{G}$ with nodes $[[1, d]]$. We assume that $\mathbf{X}$ follows a linear structural equation model $\mathbf{X} = \mathbf{B} \mathbf{X} + \mathbf{N}$, 
with $\mathbf{N} \sim \mathcal{N}(\mathbf{0}, \sigma^{2} I)$ representing independent Gaussian noise \cite{sem_ii}. Also, we consider $10$ fixed clients, each owing a private dataset upon which a DAG-GFlowNet \cite{deleu2022bayesian} is trained. Our objective is to train a GFlowNet sampling proportionally to the belief distribution defined by the product of the local rewards without directly accessing the clients' datasets, akin to \cite{Ng2022federated}. See \Cref{sec:app:dags} for further discussion. 


\noindent\textbf{Results.} \Cref{fig:dags} in \Cref{sec:app:dags} shows that the distribution learned by EP-GFlowNets over some topological features of the DAG accurately matches the target product distribution. 
Also, \Cref{fig:fed} highlights that the federated model finds structures with a significantly higher score wrt the complete data than the clients'.  
Importantly, we remark that the clients could refine the global distribution by incorporating an expert's knowledge in a personalized fashion \cite{bharti2022approximate}. However, we leave this to future works.

\subsection{Evaluating the CB loss}
\label{subsec:cb_exp}

\begin{figure}[H] 
    \centering
    \includegraphics[width=\linewidth]{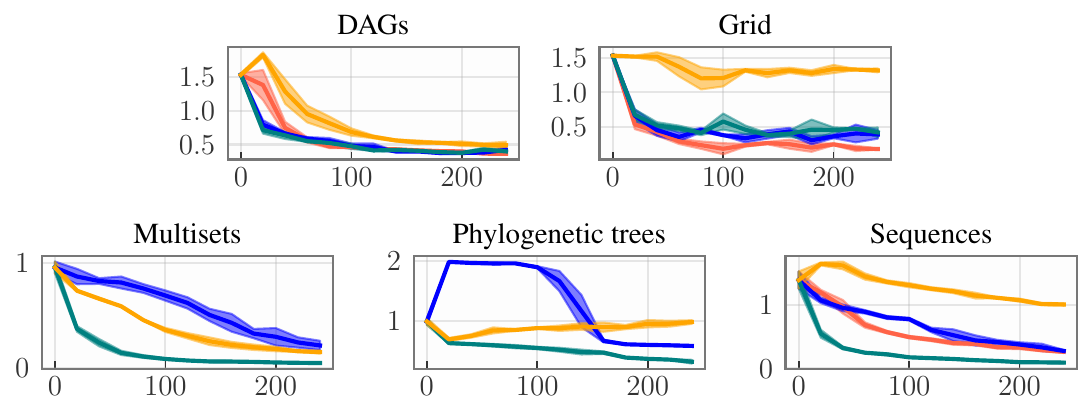}\vspace{-10pt} 
    \caption{$\textcolor{darkgreen}{\mathcal{L}_{CB}}$ performs competitively with $\textcolor{blue}{\mathcal{L}_{TB}}$, $\textcolor{orange}{\mathcal{L}_{DB}}$ and $\textcolor{red}{\mathcal{L}_{DBC}}$ in the training of conventional GFlowNets.} 
    \label{fig:contrastive}
    \vspace{-12pt} 
\end{figure}

\Cref{subsec:contrastive} presents the CB loss as a natural development given the theory of EP-GFlowNets. To evaluate its utility as a criterion to train GFlowNets in the conventional centralized (non-parallel) setting, we report the evolution during training of the $L_{1}$ error of the GFlowNet wrt the normalized reward for models trained using DB, TB, and our CB. 
\setlength{\intextsep}{0pt} 
\setlength{\columnsep}{10pt} 
We do so for all tasks in our experiments, with all GFlowNets using the same architectures for the forward and backward policies (more details in  supplement). Notably, CB led to the best convergence rate in the multiset generation, phylogeny and BNSL tasks  (\Cref{fig:contrastive}), while still performing on par with DB in the remaining domains. 
An explanation is that CB incurs a considerably simpler parametrization than DB and TB --- as we do not require estimating the flow going through each state or the target partition function. 
Indeed, when using \citet{deleu2022bayesian}'s reparametrization for DB, which obviates the estimation of state flows from state graphs where all states are terminal, implemented for grid world and sequences in \Cref{fig:contrastive}, DB and CB perform similarly. 
Moreover, to ensure the observed difference between CB and TB is not due to an insufficiently high learning rate for the log-partition function, \Cref{sec:app:moreexp} shows
results comparing the CB to TB with different lr's for $\log Z_{\phi_Z}$ in the multiset experiments (\Cref{fig:tbtomato}). Noticeably, CB outperforms TB for all rates tested. A rigorous understanding, however, of the different virtues of the diversely proposed balance conditions remains a lacking issue in the literature and is an important course of action for future research. 

\section{Conclusions} 
We proposed EP-GFlowNet as a simple and elegant solution for distributed inference over discrete distributions and federated learning of GFlowNets, which we validate on an extensive suite of experiments. Our method enjoys theoretical guarantees and builds on the concept of contrastive balance (CB). Our theoretical analysis i) guarantees correctness when local models are perfectly trained and ii) allows us to quantify the impact of errors of local models on the global one. We also observed that using CB loss led to faster convergence for the local clients when intermediate states are not terminal 
--- while being otherwise competitive. 

Remarkably, we believe EP-GFlowNets pave the way for a range of applications of distributed and federated discrete Bayesian inference. We also believe EP-GFlowNets will be useful to scale up Bayesian inference by amortizing the cost of expensive likelihood computations over different clients. In the realm of multi-objective optimization, EP-GFlowNets enable sampling from a combination of rewards ~\citep{mogfn} by leveraging pre-trained GFlowNets --- even without directly accessing the rewards. 
as recently explored by \citet{sculpting}.
We also believe the modular nature of EP-GflowNets may be instrumental in enabling the reuse and combination of large-scale models \citep{diffusions}.

\section*{Impact statement}

We proposed a framework to approximate log-pools of GFlowNets using another GFlowNet. Our method has a broad range of applications in large-scale Bayesian inference, federated learning, and multi-objective optimization. Similar to other sampling methods, our work inherits the ethical concerns involved in formulating its target distribution --- and, e.g., may propagate negative societal biases or be used to generate malicious content. On the other hand, our framework enables model reuse, which can have a positive impact on the amount of carbon emissions stemming from training large generative models. 

\section*{Acknowledgments}

Tiago da Silva, Luiz Max Carvalho, and Diego Mesquita acknowledge the support of the Funda\c{c}\~ao de Amparo \`a Pesquisa do Estado do Rio de Janeiro (FAPERJ, grant 200.151/2023), the Funda\c{c}\~ao de Amparo \`a Pesquisa do Estado de São Paulo (FAPESP, grant 2023/00815-6), and the Conselho Nacional de Desenvolvimento Científico e Tecnológico (CNPq, grant 404336/2023-0).  
Amauri Souza and Samuel Kaski were supported by the Academy of Finland (Flagship programme: Finnish Center for Artificial Intelligence FCAI), EU Horizon 2020 (European Network of AI Excellence Centres ELISE, grant agreement 951847), UKRI Turing AI World-Leading Researcher Fellowship (EP/W002973/1).
\looseness=-1 

We acknowledge Aalto Science-IT Project from Computer Science IT and FGV TIC for the provided comp. resources.
\looseness=-1 


\bibliography{references}
\bibliographystyle{icml2024}

\newpage

\appendix
\onecolumn

\appendix

\section{Proofs} \label{sec:app:p} 
\subsection{Proof of \cref{lemma:contrastive}} \label{sec:app:contrastive} 

It stems directly from the trajectory balance that, for any trajectory $\tau^\star \in \mathcal{T}$:
    \begin{alignat}{3}
         &  Z \prod_{s \rightarrow s' \in \tau^\star} {p_{F}(s \rightarrow s')} = R(x) \prod_{s \rightarrow s' \in \tau^\star} {p_{B}(s' \rightarrow s)}\\
       \iff   & Z = R(x) \prod_{s \rightarrow s' \in \tau^\star} \frac{p_{B}(s' \rightarrow s)}{p_{F}(s \rightarrow s')}
    \end{alignat}
    Therefore, applying this identity to $\tau$ and $\tau^\prime$ and equating the right-hand-sides (RHSs) yields \Cref{eq:contrastive_balance_condition}. We are left with the task of proving the converse. Note we can rewrite \Cref{eq:contrastive_balance_condition} as:
    \begin{equation}
    {R(x)} \prod_{s \rightarrow s' \in \tau} \frac{p_{B}(s' \rightarrow s)}{p_{F}(s \rightarrow s')} 
       = {R(x^\prime)} \prod_{s \rightarrow s' \in \tau'} \frac{p_{B}(s' \rightarrow s)}{p_{F}(s \rightarrow s')}. 
    \end{equation}
    If \Cref{eq:contrastive_balance_condition} holds for any pair $(\tau, \tau^\prime)$, we can vary $\tau^\prime$ freely for a fixed $\tau$ --- which implies the RHS of the above equation must be a constant with respect to $\tau^\prime$. Say this constant is $c$, then:
    \begin{alignat}{3}
       & R(x) \prod_{s \rightarrow s' \in \tau} \frac{p_{B}(s' \rightarrow s)}{p_{F}(s \rightarrow s')} = c \label{eq:lemma_balance}\\
       \iff  & R(x) \prod_{s \rightarrow s' \in \tau} {p_{B}(s' \rightarrow s)} = c \prod_{s \rightarrow s' \in \tau}{p_{F}(s \rightarrow s')},
    \end{alignat}
    and summing the above equation over all $\tau \in \mathcal{T}$ yields:
    \begin{alignat}{3}
       & \sum_{\tau \in \mathcal{T}} R(x) \prod_{s \rightarrow s' \in \tau} {p_{B}(s' \rightarrow s)} = c \sum_{\tau \in \mathcal{T}} \prod_{s \rightarrow s' \in \tau}{p_{F}(s \rightarrow s')}\\
       \implies & \sum_{\tau \in \mathcal{T}} R(x) \prod_{s \rightarrow s' \in \tau} {p_{B}(s' \rightarrow s)} = c
    \end{alignat}
    Furthermore, note that:
    \begin{alignat}{3}
        & \sum_{x \in \mathcal{X}} R(x) \sum_{\tau \in T(x)} \prod_{s \rightarrow s' \in \tau} {p_{B}(s' \rightarrow s)} = c\\
        \implies & \sum_{x \in \mathcal{X}} R(x) = c \\
        \implies & Z = c
    \end{alignat}
    Plugging $Z=c$ into \Cref{eq:lemma_balance} yields the trajectory balance condition.

\subsection{Proof of \Cref{thm:federated_condition}} \label{sec:app:federated_condition} 

The proof is based on the following reasoning. We first show that, given the satisfiability of the aggregating balance condition, the marginal distribution over the terminating states is proportional to 
\begin{equation}
    \mathbb{E}_{\tau \sim p_{B}(\cdot | x)} \left[ \prod_{1 \le i \le N} \frac{p^{(i)}_{F}(\tau)}{p^{(i)}_{B}(\tau | x)} \right],
\end{equation}
as stated in \Cref{remark1}. Then, we verify that this distribution is the same as 
\begin{equation}
    p_{T}(x) \propto \prod_{1 \le i \le N} R_{i}(x)  
\end{equation}
if the local balance conditions are satisfied. This proves the sufficiency of the aggregating balance condition for building a model that samples from the correct product distribution. The necessity follows from Proposition 16 of \citet{Foundations} and from the observation that the local balance conditions are equivalent to $\nicefrac{p_{F}^{(i)}(\tau)}{p_{B}^{(i)}(\tau | x)} = R_{i}(x)$ for each $i = 1, \dots, N$. 

Next, we provide a more detailed discussion about this proof. Similarly to \Cref{sec:app:contrastive}, notice that the contrastive nature of the aggregating balance condition implies that, if 
\begin{equation}
    \prod_{1 \le i \le N} \frac{\left(\prod_{s \rightarrow s' \in \tau} \frac{p_{F}^{(i)}(s, s')}{p_{B}^{(i)}(s', s)} \right)} {\left(\prod_{s \rightarrow s' \in \tau'} \frac{p_{F}^{(i)}(s, s')}{p_{B}^{(i)}(s', s)}\right)} = 
    \frac{\left(\prod_{s \rightarrow s' \in \tau} \frac{p_{F}(s, s')}{p_{B}(s', s)} \right)}{\left(\prod_{s \rightarrow s' \in \tau'} \frac{p_{F}(s, s')}{p_{B}(s', s)}\right)},
\end{equation}
then 
\begin{equation}
    p_{F}(\tau) = c \left(\prod_{1 \le i \le N} \frac{p_{F}^{(i)}(\tau)}{p_{B}^{(i)}(\tau | x)}\right) p_{B}(\tau | x)   
\end{equation}
for a constant $c > 0$ that does not depend either on $x$ or on $\tau$. 
Hence, the marginal distribution over a terminating state $x \in \mathcal{X}$ is 
\begin{align}
    p_{T}(x) &\coloneqq  \sum_{\tau \rightsquigarrow x} \prod_{s \rightarrow s' \in \tau} p_{F}(s \rightarrow s') \\
    &= c \sum_{\tau \rightsquigarrow x} \left(\prod_{1 \le i \le N} \frac{p_{F}^{(i)}(\tau)}{p_{B}^{(i)}(\tau | x)}\right) p_{B}(\tau | x) \\ 
    &= c \mathbb{E}_{\tau \sim p_{B}(\cdot | x)} \left[ \prod_{1 \le i \le N} \frac{p_{F}^{(i)}(\tau)}{p_{B}^{(i)}(\tau | x)} \right]. 
\end{align}
Correspondingly, $\nicefrac{p_{F}^{(i)}(\tau)}{p_{B}^{(i)}(\tau | x)} \propto R_{i}(x)$ for every $i = 1, \dots, N$ and every $\tau$ leading to $x$ due to the satisfiability of the local balance conditions. Thus, 
\begin{equation}
    p_{T}(x) \propto \mathbb{E}_{\tau \sim p_{B}(\cdot | x)} \left[ \prod_{1 \le i \le N} R_{i}(x) \right] = \prod_{1 \le i \le N} R_{i}(x), 
\end{equation}
which attests the sufficiency of the aggregating balance condition for the distributional correctness of the global model.  

\subsection{Proof of \Cref{thm:robustness}} 

Initially, recall that the Jeffrey divergence, known as the symmetrized KL divergence, is defined as 
\begin{equation}
    \mathcal{D}_{J}(p, q) = \mathcal{D}_{KL}[p || q] + \mathcal{D}_{KL}[q || p] 
\end{equation}
for any pair $p$ and $q$ of equally supported distributions. Then, let  
\begin{equation}
    \hat{\pi}(x) = \hat{Z}\, \mathbb{E}_{\tau \sim p_{B}(\cdot | x)} \left[ \prod_{1 \le i \le N} \frac{p_{F}^{(i)}(\tau)}{p_{B}^{(i)}(\tau | x)} \right] 
\end{equation}
be the marginal distribution over the terminating states of a GFlowNet satisfying the aggregating balance condition (see \Cref{remark1} and \Cref{sec:app:federated_condition}). On the one hand, notice that 
\begin{align}
    \mathcal{D}_{KL}[ \pi || \hat{\pi} ] &= \mathbb{E}_{x \sim \pi} \left[ \log \frac{\pi(x)}{\hat{\pi}(x)} \right] \\ 
    &= \mathbb{E}_{x \sim \pi} \left[ \log \pi(x) - \log Z\mathbb{E}_{\tau \sim p_{B}(\cdot | x)} \left[ \prod_{1 \le i \le N} \frac{p_{F}^{(i)}(\tau)}{p_{B}^{(i)}(\tau | x)} \right] \right] \\ 
    &= - \mathbb{E}_{x \sim \pi} \left[ \log \mathbb{E}_{\tau \sim p_{B}(\cdot | x)} \left[ \prod_{1 \le i \le N} \frac{p_{F}^{(i)}(\tau)}{p_{B}^{(i)}(\tau | x) \pi_{i}(x)} \right]  \right] - \log \hat{Z} + \log Z \\ 
    &\le - \mathbb{E}_{x \sim \pi} \left[ \log \prod_{1 \le i \le N} (1 - \alpha_{i}) \right] - \log \hat{Z} + \log Z \\ 
    &= \log \frac{Z}{\hat{Z}} + \sum_{1 \le i \le N} \log \left(\frac{1}{1 - \alpha_{i}}\right), 
\end{align}
in which $Z \coloneqq \left(\sum_{x \in \mathcal{X}} \prod_{1 \le i \le N} \pi_{i}(x)\right)^{-1}$ is $\pi$'s normalization constant. On the other hand, 
\begin{align}
    \mathcal{D}_{KL}[ \pi || \hat{\pi} ] &= \mathbb{E}_{x \sim \hat{\pi}} \left[ \log \frac{\hat{\pi}(x)}{\pi(x)} \right] \\ 
    &= \mathbb{E}_{x \sim \hat{\pi}} \left[\log Z\mathbb{E}_{\tau \sim p_{B}(\cdot | x)} \left[ \prod_{1 \le i \le N} \frac{p_{F}^{(i)}(\tau)}{p_{B}^{(i)}(\tau | x)} \right] - \log \pi(x) \right] \\ 
    &= \mathbb{E}_{x \sim \hat{\pi}} \left[ \log \mathbb{E}_{\tau \sim p_{B}(\cdot | x)} \left[ \prod_{1 \le i \le N} \frac{p_{F}^{(i)}(\tau)}{p_{B}^{(i)}(\tau | x) \pi_{i}(x)} \right]  \right] + \log \hat{Z} - \log Z \\ 
    &\le \mathbb{E}_{x \sim \hat{\pi}} \left[ \log \prod_{1 \le i \le N} (1 + \beta_{i}) \right] + \log \hat{Z} - \log Z \\ 
    &= \log \frac{\hat{Z}}{Z} + \sum_{1 \le i \le N} \log \left( 1 + \beta_{i} \right). 
\end{align}
Thus, the Jeffrey divergence between the targeted product distribution $\pi$ and the effectively learned distribution $\hat{\pi}$ is 
\begin{align}
    \mathcal{D}_{J}(\pi, \hat{\pi}) &= \mathcal{D}_{KL}[\pi || \hat{\pi}] + \mathcal{D}_{KL}[\hat{\pi} || \pi] \\ 
    &\le \log \frac{Z}{\hat{Z}} + \sum_{1 \le i \le N} \log \left( \frac{1}{1 - \alpha_{i}} \right) + \log \frac{\hat{Z}}{Z} + \sum_{1 \le i \le N} \log \left( 1 + \beta_{i} \right) \\ 
    &= \sum_{1 \le i \le N} \log \left( \frac{1 + \beta_{i}}{1 - \alpha_{i}} \right). 
\end{align}

\subsection{Proof of \Cref{prop:variational_objective}} 

We firstly recall the construction of the unbiased REINFORCE gradient estimator (Williams 1992), which was originally designed as a method to implement gradient-ascent algorithms to tackle associative tasks involving stochastic rewards in reinforcement learning. Let $p_{\theta}$ be a probability density (or mass function) differentiably parametrized by $\theta$ and $f_{\theta} \colon \mathcal{X} \rightarrow \mathbb{R}$ be a real-value function over $\mathcal{X}$ possibly dependent on $\theta$. Our goal is to estimate the gradient 
\begin{equation}
    \nabla_{\theta} \mathbb{E}_{x \sim p_{\theta}} [ f_{\theta}(x) ], 
\end{equation} 
which is not readily computable due to the dependence of $p_{\theta}$ on $\theta$. However, since 
\begin{align}
    \nabla_{\theta} \mathbb{E}_{x \sim p_{\theta}} [ f_{\theta}(x) ] &= \nabla_{\theta} \int_{x \in \mathcal{X}} f_{\theta}(x) p_{\theta}(x) \mathrm{d}x \\ 
    &= \int_{x \in \mathcal{X}} \left( (\nabla_{\theta} f_{\theta}(x)) p_{\theta}(x) \right) \mathrm{d}x + \int_{x \in \mathcal{X}} \left( (\nabla_{\theta} p_{\theta}(x)) f_{\theta}(x) \right) \mathrm{d}x \\ 
    &= \mathbb{E}_{x \sim p_{\theta}} \left[ \nabla_{\theta} f_{\theta}(x) + f_{\theta}(x) \nabla_{\theta} \log p_{\theta}(x) \right],  
\end{align}
the gradient of $f_{\theta}$'s expected value under $p_{\theta}$ may be unbiasedly estimated by averaging the quantity $\nabla_{\theta} f_{\theta}(x) + f_{\theta}(x) \nabla_{\theta} \log p_{\theta}(x)$ over samples of $p_{\theta}$. We use this identity to compute the KL divergence between the forward and backward policies of a GFlowNet. In this sense, notice that 
\begin{align} 
    \nabla_{\theta} \mathcal{D}_{KL}[p_{F} || p_{B}] &= \nabla_{\theta} \mathbb{E}_{\tau \sim p_{F}} \left[ \log \frac{p_{F}(\tau)}{p_{B}(\tau)} \right] \label{eq:gradients} \\ 
    &= \mathbb{E}_{\tau \sim p_{F}} \left[ \nabla_{\theta} \log p_{F}(\tau) + \left(\log \frac{p_{F}(\tau)}{p_{B}(\tau)}\right) \nabla_{\theta} \log p_{F} (\tau) \right] \\ 
    &= \mathbb{E}_{\tau \sim p_{F}} \left[ \left(\log \frac{p_{F}(\tau)}{p_{B}(\tau)} \right) \nabla_{\theta} \log p_{F}(\tau) \right], 
\end{align}
as $\mathbb{E}_{\tau \sim p_F} [\nabla_{\theta} \log p_{F}(\tau)] = \nabla_{\theta} \mathbb{E}_{\tau \sim p_{F}}[ 1 ] = 0$. In contrast, the gradient of the contrastive balance loss with respect to $\theta$ is 
\begin{align}
    \nabla_{\theta} \mathcal{L}_{CB}(\tau, \tau', \theta) &= \nabla_{\theta} \left( \log \frac{p_{F}(\tau)}{p_{B}(\tau)} - \log \frac{p_{F}(\tau')}{p_{B}(\tau')} \right)^{2} \\ 
    &= 2 \left( \log \frac{p_{F}(\tau)}{p_{B}(\tau)} - \log \frac{p_{F}(\tau')}{p_{B}(\tau')} \right) \left( \nabla_{\theta} \log p_{F}(\tau) - \nabla_{\theta} \log p_{F}(\tau') \right), 
\end{align}
whose expectation under the outer product distribution $p_{F} \otimes p_{F}$ equals the quantity $4 \nabla_{\theta} \mathcal{D}_{KL} [p_{F} || p_{B}]$ in \Cref{eq:gradients}. Indeed, as 
\begin{equation}
    \mathbb{E}_{\tau \sim p_{F}}\left[ \left(\log \frac{p_{F}(\tau')}{p_{B}(\tau')} \right) \nabla_{\theta} \log p_{F}(\tau) \right] = 0, 
\end{equation}
with an equivalent identity obtained by interchanging $\tau$ and $\tau'$, 
\begin{align}
    \underset{(\tau, \tau') \sim p_{F} \otimes p_{F}}{\mathbb{E}} \left[ \nabla_{\theta} \mathcal{L}_{CB} (\tau, \tau', \theta) \right] = \\
    \underset{(\tau, \tau') \sim p_{F} \otimes p_{F}}{\mathbb{E}} \left[ 2 \left( \log \frac{p_{F}(\tau)}{p_{B}(\tau)} - \log \frac{p_{F}(\tau')}{p_{B}(\tau')} \right) \left( \nabla_{\theta} \log p_{F}(\tau) - \nabla_{\theta} \log p_{F}(\tau') \right) \right] = \\ 
     \underset{ (\tau, \tau') \sim p_{F} \otimes p_{F}}{\mathbb{E}} \left[2 \left( \log \frac{p_{F}(\tau)}{p_{B}(\tau)} \right) \nabla_{\theta} \log p_{F}(\tau) + 2 \left( \log \frac{p_{F}(\tau')}{p_{B}(\tau')} \right) \nabla_{\theta} \log p_{F}(\tau') \right] = \\ 
    \underset{\tau \sim p_{F}}{\mathbb{E}} \left[ 4 \left(\log \frac{p_{F}(\tau)}{p_{B}(\tau)} \right) \nabla_{\theta} \log p_{F}(\tau) \right] = 4 \nabla_{\theta} \mathcal{D}_{KL}[p_{F} || p_{B}]. 
\end{align}
Thus, the on-policy gradient of the contrastive balance loss equals in expectation the gradient of the KL divergence between the forward and backward policies of a GFlowNet.

\section{Additional theoretical results}  
\label{sec:app:th} 

This section rigorously lays out further theoretical results which were only briefly stated in the main paper. Firstly, \autoref{sec:app:cb} (i) shows that  $\mathcal{L}_{CB}$ is, in expectation, equivalent to the \emph{variance loss} considered by \citet{robust}; and (ii) provides alternative and shorter proofs for \Cref{col:contrastive} and \Cref{prop:variational_objective} based on previously published results for $\mathcal{L}_{TB}$. Secondly, \autoref{sec:app:exp} extends our aggregation scheme to accommodate generic logarithmic pooling of GFlowNets, significantly expanding the potential applicability of EP-GFlowNets to personalized federated learning (by implementing a client-level fine-tuned log-pool \cite{xu2023personalized}) and to model composition (such as negation \cite{diffusions, sculpting}).

\subsection{Relationship of $\mathcal{L}_{CB}$ to other losses} 
\label{sec:app:cb} 

\noindent\textbf{Alternative proofs for \Cref{col:contrastive} and \Cref{prop:variational_objective}.} To start with, we define $\mathcal{V}_{TB}(\tau) = \log p_{F}(\tau) + \log Z - \log p_{B}(\tau | x_{\tau}) - \log R(x_{\tau})$ for the signed violation to the TB condition in log-space; $x_{\tau}$ represents the terminal state of the complete trajectory $\tau$. Note then that  
\begin{equation} 
\mathcal{L}_{CB}(\tau, \tau') = \left(\mathcal{V}_{TB}(\tau) - \mathcal{V}_{TB}(\tau')\right)^{2} \text{ and } \mathcal{L}_{TB}(\tau) = \mathcal{V}_{TB}(\tau)^{2}.  
\end{equation} 
In this context, Proposition 1 of \cite{malkin2022trajectory} states the sufficiency of the condition $\mathcal{V}_{TB}(\tau)^{2} = 0$ --- and equivalently of $\mathcal{V}_{TB}(\tau) = 0$ --- for each $\tau$ to ensure that the GFlowNets' policies sample proportionally to the reward distribution $R(x)$. As a consequence, $\mathcal{L}_{CB}(\tau, \tau') = 0$ ensures that $\mathcal{V}_{TB}$ is constant and, in particular, that there exists a $c > 0$ such that 
\begin{equation}
    \mathcal{V}_{TB} + (\log c - \log Z) = \log p_{F}(\tau) + \log c - \log p_{B}(\tau | x_{\tau}) - \log R(x_{\tau}) = 0. 
\end{equation}
This condition corresponds to a reparametrization of the TB and entails, by Proposition 1 of \cite{malkin2022trajectory}, \Cref{col:contrastive}. Moreover, since 
\begin{equation}
    \mathbb{E}_{(\tau, \tau') \sim p_{F} \otimes p_{F}} \left[ \nabla_{\theta_{F}} \mathcal{L}_{CB}(\tau, \tau') \right] = \mathbb{E}_{(\tau, \tau') \sim p_{F} \otimes p_{F}} \left[ 2\left(\mathcal{V}_{TB}(\tau) - \mathcal{V}_{TB}(\tau')\right) \nabla_{\theta_{F}} \left( \mathcal{V}_{TB}(\tau) - \mathcal{V}_{TB}(\tau')\right) \right] 
\end{equation}
by the chain rule, 
\begin{equation}
    \mathbb{E}_{\tau \sim p_{F}} \left[ \nabla_{\theta_{F}} \mathcal{L}_{TB}(\tau) \right] = \mathbb{E}_{\tau \sim p_{F}} \left[ 2 \mathcal{V}_{TB}(\tau) \nabla_{\theta_{F}} \mathcal{V}_{TB}(\tau) \right] = 2 \mathcal{D}_{KL}\left[ p_{F} || p_{B} \right] 
\end{equation}
by Proposition 1 of \cite{malkin2023gflownets}, and 
\begin{equation}
    \mathbb{E}_{\tau \sim p_{F}}\left[ \nabla_{\theta_{F}} \mathcal{V}_{TB}(\tau) \right] = \sum_{\tau} p_{F}(\tau) \nabla_{\theta_{F}} \log p_{F}(\tau) = \nabla_{\theta_{F}} \sum_{\tau} p_{F}(\tau) = 0, 
\end{equation}
one may conclude that 
\begin{equation}
    \mathbb{E}_{(\tau, \tau') \sim p_{F} \otimes p_{F}} \left[ \mathcal{L}_{CB} (\tau, \tau') \right] = 4 \mathcal{D}_{KL}\left[p_{F} || p_{B}\right], 
\end{equation}
thereby showing \autoref{prop:variational_objective}. 

\noindent\textbf{Relationship of $\mathcal{L}_{CB}$ to $\mathcal{L}_{VL}$.} \citet{robust} proposed to train a GFlowNet by minimizing an expectation of the \textit{variance loss}, 
\begin{equation}
    \mathcal{L}_{VL}(\tau) = \left(\mathcal{V}_{TB}(\tau) - \mathbb{E}_{\tau'}[\mathcal{V}_{TB}(\tau')]\right)^{2}; 
\end{equation}
in practice, the inner expectation was locally estimated using the average violation of batch of trajectories. Notably, the CB loss, derived from our proposed contrastive balance condition, equals twice $\mathcal{L}_{VL}$ in expectation --- when the training policy samples the trajectories independently, since 
\begin{equation}
    \mathbb{E}_{\tau, \tau'} \left[ \left(\mathcal{V}_{TB}(\tau) - \mathcal{V}_{TB}(\tau')\right)^{2} \right] = 2 \mathbb{E}_{\tau}\left[ \left( \mathcal{V}_{TB}(\tau) - \mathbb{E}_{\tau'} \left[ \mathcal{V}_{TB}(\tau') \right]\right)^{2} \right] = 2 \mathbb{E}_{\tau} \left[ \mathcal{L}_{VL}(\tau) \right].  
\end{equation}
Remarkably, this equation may also be elegantly deduced from the general fact that the expectation of the squared difference between two i.i.d. random variables equals twice their variance. 

\noindent\textbf{$\mathcal{L}_{CB}$ and VI when $p_{B}$ is parameterized.} The theorem below shows that the CB loss coincides with the KL divergence between the forward and backward policies in terms of expected gradients when $p_{F}$ and $p_{B}$ are disjointly parameterized. 

\begin{theorem} 
    Let $\phi_{B}$ represent the parameters of the backward policy. Also, let $p_{B} \otimes p_{B}$ be the outer-product distribution assigning probability $\frac{1}{Z^{2}} R(x) R(x’) p_{B}(\tau | x) p_{B}(\tau | x’)$ to each pair of trajectories $(\tau, \tau’)$ terminating respectively at $x$ and $x’$. Then, the CB loss satisfies 
    \begin{equation} 
        \mathbb{E}_{\tau, \tau’ \sim p_{B} \otimes p_{B}} \left[ \nabla_{\phi_{B}} \mathcal{L}_{CB} (\tau, \tau’ ; \phi_{B}) \right] = \frac{1}{4} \mathcal{D}_{KL} \left[ p_{B} || p_{F} \right].
    \end{equation} 
\end{theorem} 

Importantly, the parameterization of $p_{B}$ doubles the number of forward and backward passes performed in training, which often obfuscates the advantages enacted by the improved credit assignment provided by an appropriate $p_{B}$ \cite{towards}. Hence, while acknowledging the importance of designing effective learning schemes for $p_{B}$, we fix the backward policy as an uniform. 

\subsection{Exponentially weighted distributions} 
\label{sec:app:exp} 

This section extends our theoretical results and shows how to train an EP-GFlowNet to sample from a logarithmic pool of locally trained GFlowNets. 
Henceforth, let $R_{1}, \dots, R_{N} \colon \mathcal{X} \rightarrow \mathbb{R}_{+}$ be non-negative functions over $\mathcal{X}$ and assume that each client $n = 1, \dots, N$ trains a GFlowNet to sample proportionally to $R_{n}$. The next propositions show how to train a GFlowNet to sample proportionally to an exponentially weighted distribution $\prod_{n=1}^{N} R_{n}(x)^{\omega_{n}}$ for non-negative weights $\omega_{1}, \dots, \omega_{N}$.   
We omit the proofs since they are essentially identical to the ones presented in \Cref{sec:app:p}.  

Firstly, \Cref{thmbis:federated_condition} below proposes a modified balance condition for the global GFlowNet and shows that the satisfiability of this condition leads to a generative model that samples proportionally to the exponentially weighted distribution.  

\begin{thmbis}{thm:federated_condition}[Aggregating balance condition] \label{thmbis:federated_condition} 
Let $\left(p_F^{(1)}, p_F^{(1)}\right),  \dots, \left(p_F^{(N)}, p_F^{(N)}\right): V^2 \rightarrow \mathbb{R}^+$ be pairs of forward and backward policies from $N$ GFlowNets sampling respectively proportional to $R_1, \ldots, R_N  : \mathcal{X} \rightarrow \mathbb{R}^+$.
    Then, another GFlowNet with forward and backward policies $p_F, p_B \in V^2 \rightarrow \mathbb{R}^+$ samples proportionally to $R(x) \defeq \prod_{n=1}^{N} R(x)^{\omega_{n}}$ if and only if the following condition holds for any terminal  trajectories $\tau, \tau^\prime \in \mathcal{T}$:
    \begin{equation}
        \prod_{1 \le i \le N} \frac{\left(\prod_{s \rightarrow s' \in \tau} \frac{p_{F}^{(i)}(s, s')}{p_{B}^{(i)}(s', s)} \right)^{\omega_{i}}} {\left(\prod_{s \rightarrow s' \in \tau'} \frac{p_{F}^{(i)}(s, s')}{p_{B}^{(i)}(s', s)}\right)^{\omega_{i}}} = 
        \frac{\left(\prod_{s \rightarrow s' \in \tau} \frac{p_{F}(s, s')}{p_{B}(s', s)} \right)}{\left(\prod_{s \rightarrow s' \in \tau'} \frac{p_{F}(s, s')}{p_{B}(s', s)}\right)}. 
        \label{eq:app:fed_balance}
    \end{equation}
\end{thmbis}

Secondly, \Cref{thmbis:robustness} provides an upper bound on the discrepancy between the targeted and the learned global distributions under controlled local errors --- when the local distributions are heterogeneously pooled. Notably, it suggests that the effect of the local failures over the global approximation may be mitigated by reducing the weights associated with improperly trained local models.  

\begin{thmbis}{thm:robustness}[Influence of local failures] \label{thmbis:robustness} 
   Let $\pi_n \defeq R_{n}/Z_{n}$ and $p_{F}^{(n)}$ and $p_{B}^{(n)}$ be the forward and backward policies of the $n$th client. We use $\tau \rightsquigarrow x$ to indicate that $\tau \in \mathcal{T}$ is finished by $x \rightarrow s_{f}$. Suppose that the local balance conditions are lower- and upper-bounded $\forall\, n=1,\ldots,N$ as per 
   \begin{equation}
       1 - \alpha_{n} \le \min_{x \in \mathcal{X}, \tau \rightsquigarrow x} \frac{p_{F}^{(n)}(\tau)}{p_{B}^{(n)}(\tau | x) \pi_{n}(x)} \le \max_{x \in \mathcal{X}, \tau \rightsquigarrow x} \frac{p_{F}^{(n)}(\tau)}{p_{B}^{(n)}(\tau | x) \pi_{n}(x)} \le 1 + \beta_{n} 
   \end{equation}
    where $\alpha_{n} \in (0, 1)$ and $\beta_{n} > 0$. The Jeffrey divergence $\mathcal{D}_{J}$ between the global model $\hat{\pi}(x)$ that fulfills the aggregating balance condition in \Cref{eq:app:fed_balance} and $\pi(x) \propto \prod_{n=1}^{N} \pi_{n}(x)^{\omega_{n}}$ then satisfies 
    \begin{equation}
        \mathcal{D}_{J}(\pi, \hat{\pi}) \le \sum_{n=1}^{N} \omega_{n} \log\left(\frac{1 + \beta_{n}}{1 - \alpha_{n}}\right). \label{eq:app:jeff_bound} 
    \end{equation}
\end{thmbis}

Interestingly, one could train a \textit{conditional} GFlowNet \citep{Bengio2021, robust} to build an amortized generative model able to sample proportionally to $\prod_{n=1}^{N} R_{n}(x)^{\omega_{n}}$ for any non-negative weights $(\omega_{1}, \dots, \omega_{N})$ within a prescribed set. This is a promising venue for future research. 



\section{Additional experiments and implementation details} 

\begin{algorithm}[!t] 
    \caption{Training of EP-GFlowNets}\label{alg:cap}
    \begin{algorithmic}
    \Require $\left(p_{F}^{(1)}, p_{B}^{(1)}\right), \dots, \left(p_{F}^{(K)}, p_{B}^{(K)}\right)$ clients' policies, $R_{1}, \dots, R_{K}$ clients' rewards, $\left(p_{F}, p_{B}\right)$ parameterized global policies, $E$ number of epochs for training, $u_{F}$ uniform policy 
    \Ensure $p^{\intercal}(x) \propto R(x) \coloneq \prod_{1 \le k \le K} R_{k}(x)$ 
    \ParFor{$k \in \{1, \dots, K\}$} \Comment{Train the clients' models in parallel}
        \State train the policies \, $\left(p_{F}^{(k)}, p_{B}^{(k)}\right)$ to sample proportionally to $R_{k}$
    \EndParFor

    \For{$e \in \{1, \dots, E\}$} \Comment{Train the global model} 
    
        \State $\mathcal{B} \gets \left\{(\tau, \tau') \colon \tau, \tau' \sim \nicefrac{1}{2} \cdot p_{F} + \nicefrac{1}{2} \cdot u_{F}\right\}$ \Comment{Sample a batch of trajectories} 
        \State $L \gets \frac{1}{|\mathcal{B}|} \sum_{\tau, \tau' \in \mathcal{B}} \mathcal{L}_{AB}\left(\tau, \tau' ; \left\{\left(p_{F}^{(1)}, p_{B}^{(1)}\right), \dots, \left(p_{F}^{(K)}, p_{B}^{(K)}\right) \right\}\right)$ 
        \State Update the parameters of $p_{F}$ and $p_{B}$ through gradient descent on $L$ 
    \EndFor 

    \end{algorithmic}
\end{algorithm}

This section is organized as follows. First, \Cref{sec:app:experiments} describes the experimental setup underlying the empirical evaluation of EP-GFlowNets in \Cref{sec:experiments}. Second, \Cref{sec:app:pcvi} exhibits the details of the variational approximations to the combined distributions used as baselines in \Cref{tab:gflownets}. Third, \Cref{sec:app:criteria} specifies our settings for comparing the training convergence speed of different optimization objectives. 
\Cref{alg:cap} illustrates the training procedure of EP-GFlowNets. The computer code for reproducing our experiments will be publicly released at \href{https://github.com/ML-FGV/ep-gflownets}{github.com/ML-FGV/ep-gflownets}. 

\vspace{-6pt} 

\subsection{Experimental setup} \label{sec:app:experiments}  

In the following, we applied the same optimization settings for each environment. For the stochastic optimization, we minimized the contrastive balance objective using the AdamW optimizer \citep{loshchilov2018decoupled} for both local and global GFlowNets. We trained the models for $5000$  epochs (20000 for the grid world) with a learning rate equal to $3 \cdot 10^{-3}$ with a batch size dependent upon the environment. Correspondingly, we define the $L_{1}$ error between the distributions $\pi$ and $\hat{\pi}$ as two times the total variation distance between them, $\|\pi - \hat{\pi}\|_{1} \coloneqq \sum_{x \in \mathcal{X}} |\pi(x) - \hat{\pi}(x)|$. 
For the grid world, design of sequences and federated BNSL setups, all intermediate GFlowNet states are also terminal, since they are connected to a sink state. For the remaining setups, most states are not terminal and exist solely as intermediate steps between the initial state and the target distribution's support.

\paragraph{Grid world.} We considered a two-dimensional grid with length size $9$ as the environment for the results of both \Cref{tab:gflownets} and \Cref{fig:grid}. To parameterize the forward policy, we used an MLP with two 64-dimensional layers and a LeakyReLU activation function between them \citep{maas2013rectifier}. For inference, we simulated $10^{6}$ environments to (i) compute the $L_{1}$ error between the targeted and the learned distributions; and (ii) selected the $800$ most rewarding samples. We utilized a batch size equal to 1024 during both the training and inference phases. 

\paragraph{Design of sequences.} We trained the GFlowNets to generate sequences of size up to $6$ with elements selected from a set of size $6$. We parametrized the forward policies with a single 64-dimensional layer bidirectional LSTM network followed by an MLP with two 64-dimensional layers \citep{graves2012long}. For training, we used a batch size of 512. For inference, we increased the batch size to 1024 and we sampled $10^{6}$ sequences to estimate the quantities reported in \Cref{tab:gflownets} and \Cref{fig:sequence}. 

\paragraph{Multiset generation.} We designed the GFlowNet to generate multisets of size $8$ by iteratively selecting elements from a set $U$ of size $10$. Moreover, we endowed each element within $U$ with a learnable and randomly initialized $10$-dimensional embedding. To estimate the transition probabilities at a given state $s$, we applied an MLP with two 64-dimensional layers to the sum of the embeddings of the elements in $s$. During training, we used a batch size of 512 to parallely generate multiple multisets and reduce the noiseness of the backpropagated gradients. During inference, we increased the batch size to 1024 and generated $10^{6}$ samples to generate the results reported in \Cref{tab:gflownets} and \Cref{fig:multiscale}. 

\begin{figure}
    \centering
    \includegraphics[width=.8\textwidth]{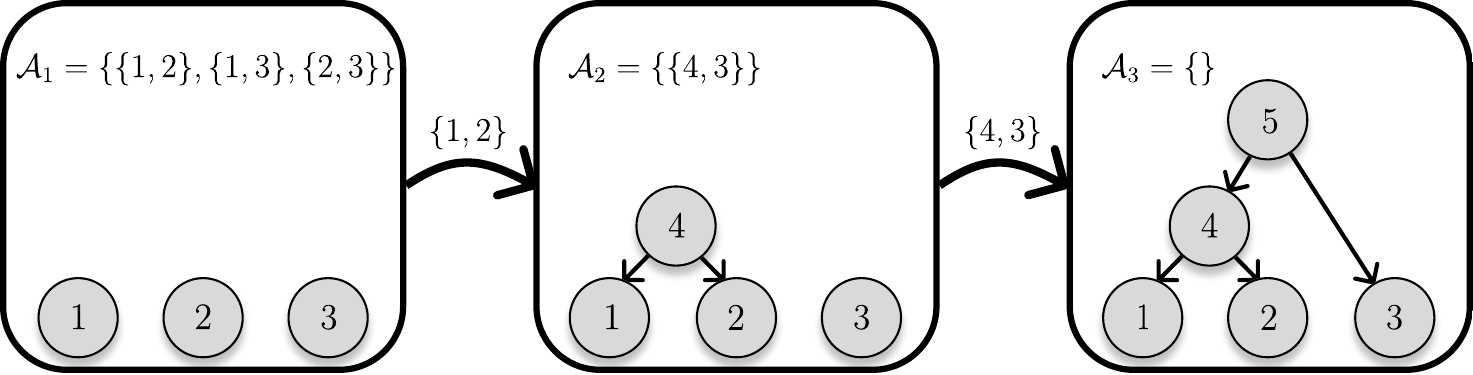}
    \caption{\textbf{An illustration of the generative process for phylogenetic trees' topologies.} We iteratively select two trees to join their roots. The final state corresponds to a single, connected graph.}
    \label{fig:trees}
\end{figure}

\paragraph{Bayesian phylogenetic inference.} We devised a GFlowNet to learn a posterior distribution over the space of rooted phylogenetic trees with $7$ leaves and fixed branch lengths. Each state is represented as a forest. Initially, each leaf belongs to a different singleton tree. An action consists of picking two trees and joining their roots to a newly added node. The generative process is finished when all nodes are connected in a single tree (see \Cref{fig:trees}; a similar modeling was recently considered by \cite{zhou2023phylogfn}).To estimate the policies at the (possibly partially built) tree $t$, we used a graph isomorphism network \citep[GIN;][]{xu2018powerful} with two 64-dimensional layers to generate node-level representations for $t$ and then used an MLP to project the sum of these representations to a probability distribution over the viable transitions at $t$. We used a tempered version of the likelihood to increase the sparsity of the targeted posterior. Importantly, we selected a batch size of 512 for training and of 1024 for inference. Results for \Cref{tab:gflownets} and \Cref{fig:phylo} are estimates based on $10^{5}$ trees drawn from the learned distributions. To evaluate the likelihood function, we considered Jukes \& Cantor's nucleotide substitution model for the observed sites \cite{jukes}, which assigns the same instantaneous substitution rate for each pair of nucleotides. 

\paragraph{Bayesian network structure learning.} Each client trains a DAG-GFlowNet \citet{deleu2022bayesian} to sample proportionally to the reward function evaluated at a small dataset of $20$ points. However, in contrast \citet{deleu2022bayesian}, which implemented a linear transformer for parameterizing GFlowNet's forward policy, we use a 2-layer 128-dimensional MLP; the backward policy is fixed as an uniform distribution over a state's parents. For the global model, we increase the MLP's latent layers' size to 256 units each. To estimate the probability $\mathbb{P}_{G}[U \rightarrow V]$, we sample $\{G_{1}, \ldots, G_{N}\}$ and compute 
\begin{equation}
    \frac{1}{N} \sum_{1 \le n \le N} \mathbbm{1}[U \rightarrow V | G_{n}], 
\end{equation}
with $\mathbbm{1}[U \rightarrow V | G_{n}]$ indicating whether the edge $U \rightarrow V$ exists in $G_{n}$. We compare this quantity with the nominal value obtained by enumerating the target distribution $\pi$ support and directly evaluating the expectation $\mathbb{E}_{G \sim \pi} \left[ \mathbbm{1}[U \rightarrow V | G] \right]$, similarly to \cite{deleu2022bayesian}. 

Our implementations were based on \texttt{PyTorch} \citep{Paszke_PyTorch_An_Imperative_2019} and on \texttt{PyTorch Geometric} \citep{Fey/Lenssen/2019}. 

\subsection{Parallel Categorical Variational Inference} \label{sec:app:pcvi} 

As a simplistic approach to combining the locally learned distributions over compositional objects, we variationally approximate them as the product of categorical distributions over the objects' components. For this, we select the parameters that minimize the reverse Kullback-Leibler divergence between the GFlowNet's distribution $p_{T}$ and the variational family $\mathcal{Q}$, 
\begin{equation} \label{eq:variational} 
    \hat{q} = \argmin_{q \in \mathcal{Q}} \text{KL}[p_{T} || q] = \argmin_{q \in \mathcal{Q}} - \mathbb{E}_{x \sim p_{T}} [ \log q(x) ], 
\end{equation}
which, in asymptotic terms, is equivalent to choosing the parameters that maximize the likelihood of the GFlowNet's samples under the variational model. Then, we use a logarithmic pool of these local variational approximations as a proxy for the global model. In the next paragraphs, we present the specific instantiations of this method for the domains we considered throughout our experiments. We used the same experimental setup of \Cref{sec:app:experiments} to train the local GFlowNets. 

\paragraph{Grid world.} An object in this domain is composed of its two coordinates in the grid. For a grid of width $W$ and height $H$, we consider the variational family
\begin{equation}
    \mathcal{Q} = \{(\phi, \psi) \in \Delta^{W + 1} \times \Delta^{H + 1} \colon q_{\phi, \psi}(x, y) = \text{Cat}(x | \phi) \text{Cat}(y | \psi)\}, 
\end{equation}
in which $\Delta^{d}$ is the $d$-dimensional simplex and $\text{Cat}(\phi)$ ($\text{Cat}(\psi)$) is a categorical distribution over $\{0, \dots, W\}$ ($\{0, \dots, H\}$) parameterized by $\phi$ ($\psi$). Then, given the $N$ variational approximations $\left(q_{\phi^{(1)}, \psi^{(1)}}\right), \dots, \left(q_{\phi^{(N)}, \psi^{(N)}}\right)$ individually adjusted to the distributions learned by the local GFlowNets, we estimate the unnormalized parameters $\tilde{\phi}$ and $\tilde{\psi}$ of the variational approximation to the global distribution over the positions within the grid as 
\begin{equation}
    \tilde{\phi} = \bigodot_{1 \le i \le N} \phi^{(i)} \text{ and } \tilde{\psi} = \bigodot_{1 \le i \le N} \psi^{(i)}. 
\end{equation}
Then, we let $\phi = \nicefrac{\phi_{u}}{\phi_{u}^{\intercal} \mathbf{1}_{W + 1}}$ and $\psi = \nicefrac{\psi_{u}}{\psi_{u}^{\intercal} \mathbf{1}_{H + 1}}$, with $\mathbf{1}_{d}$ as the d-dimensional vector of $1$s, be the parameters of the global model.  

\paragraph{Design of sequences.} We represent sequences of size up to $T$ over a dictionary $V$ as a tuple $(S, (x_{1}, \dots, x_{S}))$ denoting its size $S$ and the particular arrangement of its elements $(x_{1}, \dots, x_{S})$. This is inherently modeled as a hierarchical model of categorical distributions, 
\begin{align}
    S \sim \text{Cat}(\theta), \\ 
    x_{i} \sim \text{Cat}(\phi_{i, S} | S) \text{ for } i \in \{1, \dots, S\},  
\end{align}
which is parameterized by $\theta \in \Delta^{T}$ and $\phi_{\cdot, S} \in \mathbb{R}^{S \times |V|}$ for $S \in \{1, \dots, T\}$. We define our family of variational approximations as the collection of all such hierarchical models and estimate the parameters $\theta$ and $\phi$ accordingly to \Cref{eq:variational}. In this case, let $\left(\theta^{(1)}, \phi^{(1)}\right), \dots, \left(\theta^{(N)}, \phi^{(N)}\right)$ be the parameters associated with the variational approximations to each of the $N$ locally trained GFlowNets. The unnormalized parameters $\tilde{\theta}$ and $\tilde{\phi}$ of the combined model that approximates the global distribution over the space of sequences are then 
\begin{equation}
    \tilde{\theta} = \bigodot_{1 \le i \le N} \theta^{(i)} \text{ and } \tilde{\phi}_{\cdot, S} = \bigodot_{1 \le i \le N} \phi_{\cdot, S}^{(i)} \text{ for } S \in \{1, \dots, T\}, 
\end{equation}
whereas the normalized ones are $\theta = \nicefrac{\tilde{\theta}}{\tilde{\theta}^{\intercal} \mathbf{1}_{T}}$ and $\phi_{\cdot, S} = \text{diag}(\tilde{\phi}_{\cdot, S} \mathbf{1}_{|V|})^{-1} \tilde{\phi}_{\cdot, S}$. 

\paragraph{Multiset generation.} We model a multiset $\mathcal{S}$ of size $S$ as a collection of independently sampled elements from a warehouse $\mathcal{W}$ with replacement. This characterizes the variational family 
\begin{equation}
    \mathcal{Q} = \left\{ q(\cdot | \phi) \colon q(\mathcal{S} | \phi) = \prod_{s \in \mathcal{S}} \text{Cat}(s | \phi) \right\}, 
\end{equation}
in which $\phi$ is the parameter of the categorical distribution over $\mathcal{W}$ estimated through \Cref{eq:variational}. Denote by $\phi^{(1)}, \dots, \phi^{(N)}$ the estimated parameters that disjointly approximate the distribution of $N$ locally trained GFlowNets. We then variationally approximate the logarithmically pooled global distribution as $q(\cdot | \phi) \in \mathcal{Q}$ with $\phi = \nicefrac{\tilde{\phi}}{\tilde{\phi}^{\intercal}\mathbf{1}_{|\mathcal{W}|}}$, in which 
\begin{equation}
    \tilde{\phi} = \bigodot_{1 \le i \le N} \phi^{(i)}. 
\end{equation}
Notably, the best known methods for carrying out Bayesian inference over the space of phylogenetic trees are either based on Bayesian networks \citep{matsen} or MCMC, neither of which are amenable to data parallelization and decentralized distributional approximations without specifically tailored heuristics. More precisely, the product of Bayesian networks may not be efficiently representable as a Bayesian network, and it is usually not possible to build a global Markov chain whose stationary distribution matches the product of the stationary distributions of local Markov chains. Moreover, any categorical variational approximation factorizable over the trees' clades would not be correctly supported on the space of complete binary trees and would lead to frequently sampled invalid graphs. By a similar reasoning, we do not compare EP-GFlowNets to a composition of tractable variational approximations to the local models, such as \cite{geffner2022deep, zheng2018dags}, in the problem of federated Bayesian structure learning.   

\subsection{Comparison of different training criteria} \label{sec:app:criteria} 

\paragraph{Experimental setup.} We considered the same environments and used the same neural network architectures described in \Cref{sec:app:experiments} to parametrize the transition policies of the GFlowNets.
Importantly, the implementation of the DB constraint and of the FL-GFlowNet requires the choice of a parametrization for the state flows \citep{Foundations, LingTrajectory}. We model them as an neural network with an architecture that essentially mirrors that of the transition policies --- with the only difference being the output dimension, which we set to one. Moreover, we followed suggestions in \citep{LingTrajectory, malkin2022trajectory} and utilized a learning rate of $3 \cdot 10^{-3}$ for all parameters of the policy networks except for the partition function's logarithm $\log Z$ composing the TB constraint, for which we used a learning rate of $1 \cdot 10^{-1}$. Noticeably, we found that this heterogeneous learning rate scheme is crucial to enable the training convergence under the TB constraint.    

\paragraph{Further remarks regarding \Cref{fig:contrastive}.} In \Cref{fig:contrastive}, we observed that $\mathcal{L}_{CB}$ and $\mathcal{L}_{TB}$ perform similarly in the grid world and in design of sequences tasks. A reasonable explanation for this is that such criteria are identically parameterized in such domains, as $\mathcal{L}_{DB}$ reduces to $R(s') p_{B}(s | s') p_{F}(s_{f} | s) = R(s) p_{F}(s' | s) p_{F}(s_{f} | s')$ in environments where every state is terminal \cite{deleu2022bayesian}. Thus, $F$ vanishes and hence the difficult estimation of this function is avoided.  

\subsection{Additional experiments}
\label{sec:app:moreexp} 

\begin{table}[t!] 
    \caption{\textbf{Quality of the parallel approximation}. The global model's performance does not critically depend on the clients' training objective; it relies only on the goodness-of-fit of their models.}
    \label{tab:tbtomato}
    \centering
    \centering
    \begin{tabular}{c c cc cc c}
             \toprule 
         & \multicolumn{2}{c}{\textbf{Grid World}} & \multicolumn{2}{c}{\textbf{Multisets}} & \multicolumn{2}{c}{\textbf{Sequences}} \\ 
         & $L_{1} \downarrow$ & Top-800 $\uparrow$ & $L_{1} \downarrow$ & Top-800 $\uparrow$ & $L_{1} \downarrow$ & Top-800 $\uparrow$ \\ 
         \midrule 
         \multirow{2}{*}{EP-GFlowNet (CB)} & 

        $0.038$ 
         & $-6.355$ 

         & $0.130$
         & $27.422$ 

         & $0.005$
         & $-1.535$ \\ 

         & $\textcolor{gray}{ {\scriptstyle( \pm 0.016) } }$
         & $\textcolor{gray}{ {\scriptstyle( \pm 0.000) } }$

         & $\textcolor{gray}{ {\scriptstyle( \pm 0.004) } }$
         & $\textcolor{gray}{ {\scriptstyle( \pm 0.000) } }$ 

         & $\textcolor{gray}{ {\scriptstyle( \pm 0.002) } }$
         & $\textcolor{gray}{ {\scriptstyle( \pm 0.000) } }$ \\ 

         \multirow{2}{*}{EP-GFlowNets (TB)} & 

         $0.039$   
         & $-6.355$ 

         & $0.131$ 
         & $27.422$ 

         & $0.006$
         & $-1.535$ \\ 

         & $\textcolor{gray}{ {\scriptstyle( \pm 0.006) } }$
         & $\textcolor{gray}{ {\scriptstyle( \pm 0.000) } }$

         & $\textcolor{gray}{ {\scriptstyle( \pm 0.018) } }$
         & $\textcolor{gray}{ {\scriptstyle( \pm 0.000) } }$

         & $\textcolor{gray}{ {\scriptstyle( \pm 0.005) } }$ 
         & $\textcolor{gray}{ {\scriptstyle( \pm 0.000) } }$ \\
        \bottomrule 
    \end{tabular}
\end{table}

\begin{figure}
    \centering
    \includegraphics[width=.8\textwidth]{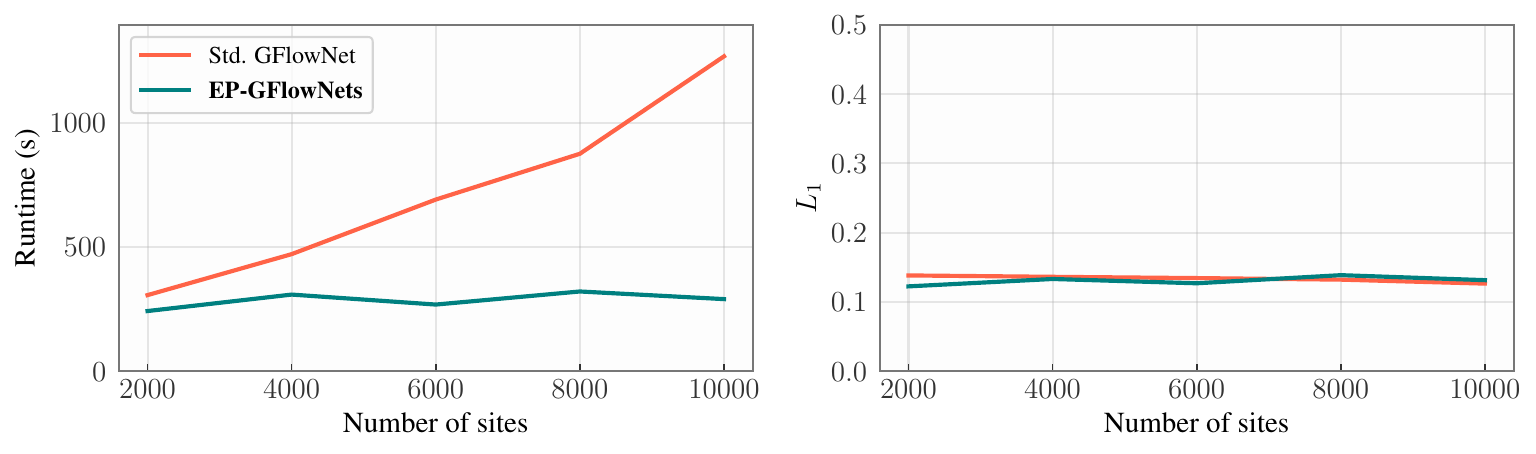}
    \caption{\textbf{EP-GFlowNets achieve a significant reduction in runtime relatively to a standard GFlowNet} in the task of Bayesian phylogenetic inference. The left plot highligths that the training time for an EP-GFlowNet remains approximately constant when an increasing set of observed samples is equally partitioned between a correspondingly increasing set of clients, each receiving $1000$ sites, whereas the training time of a standard GFlowNet grows roughly linearly on the number of samples. Importantly, the right plot shows the asynchronously trained  model performs comparably to the synchronously trained one for all the considered data sizes.}
    \label{fig:federated}
\end{figure}

\paragraph{Reduction in runtime achieved due to a distributed formulation.} \Cref{fig:federated} shows that our distributed inference framework enacts a significant reduction in runtime in the task of Bayesian phylogenetic inference relatively to a centralized approach --- while maintaining a competitive performance under the $L_{1}$ metric\footnote{Note that, in \Cref{fig:federated}, the runtime for EP-GFlowNet equals the maximum time for training the clients' models plus the running time of our aggregation step.}. This underlines the effectiveness of our method for Bayesian inference when the target distribution is computationally difficult to evaluate due to an expensive-to-compute likelihood function (which, for phylogenies, requires running Felsenstein's dynamic programming algorithm \cite{Felsenstein1981} to carry out message-passing in a graphical model). In this context, EP-GFlowNets may be also of use in Bayesian language modeling with large language models (LLMs), in which case the likelihood function is provided by a LLM with a notoriously high computational footprint; see \cite{hu2023amortizing}. For this experiment, we considered the same setup described in \Cref{sec:app:experiments}, partitioning the data among an increasing number of $\{2, 4, 6, 8, 10\}$ identical clients, each processing a set of $1000$ random sites, and training the corresponding centralized and distributed models to sample from the full posterior. The reported times were measured in a high-performance Linux machine equipped with an AMD EPYC 7V13 64-core processor, 216 GB DDR4 RAM and a NVIDIA\textsuperscript{TM} A100 80 GB PCIe 4.0 GPU. Notably, we found out that training a GFlowNet in a CPU is considerably faster in this case than training it in a GPU --- possibly due to the iterative nature of the generative process and the relatively small size of the policy networks, whose accelerated evaluation in a GPU does not compensate for the corresponding slow-down of the tree-building process. 

\paragraph{Comparison between TB and CB with different learning rates.} 
\begin{wrapfigure}[9]{r}{.45\textwidth}
    \centering
    \includegraphics[width=.8\linewidth]{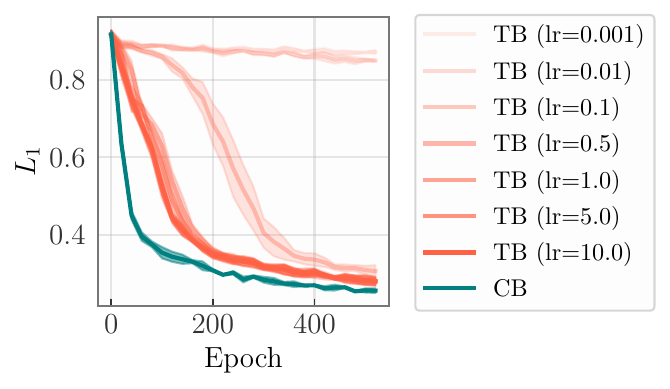}
    \vspace{-12pt} 
    \caption{CB outperforms TB for different \emph{lr}'s for $\log Z$.}
    \label{fig:tbtomato}
\end{wrapfigure}
\Cref{fig:tbtomato} shows that increasing the learning rate for $\log Z_{\phi_{Z}}$ significantly accelerates the training convergence for the TB objective. In this experiment, the learning rate for the other parameters was fixed at $10^{-3}$ --- following the setup of \citet[Appendix B]{malkin2022trajectory}. 
However, CB leads to faster convergence relatively to TB for all considered learning rates. In practice, though, note that finding an adequate learning rate for $\log Z_{\phi_{Z}}$ may be a very difficult and computationally exhaustive endeavor that is completely avoided by implementing the CB loss.

\paragraph{Sampling from product of distributions.} 
\begin{wrapfigure}{r}{.45\textwidth} 
    \centering
    \vspace{-6pt}
    \includegraphics[width=.5\linewidth]{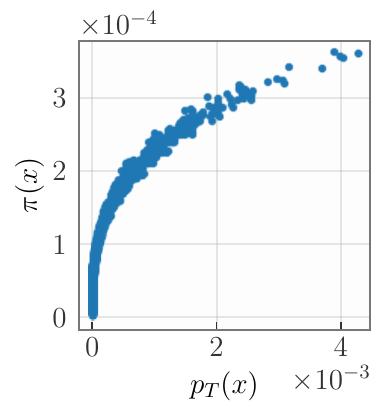}
    \vspace{-6pt} 
    \caption{Sampling from the product of GFlowNets' policies isn't equivalent to sampling from the local targets' product.}
    \label{fig:app:pp} 
\end{wrapfigure}
As a first approximation, one could employ a simple rejection sampling procedure to sample from a product of discrete distributions $q_{1}, \ldots, q_{K}$ over a shared support: draw independently a $x_{i}$ from $q_{i}$ and accept the resulting $x_{i}$ only if it was sampled by every other model. Nonetheless, this approach scales very badly as one increases either the number of distributions or the size of their common support. \citet{HintonProduct}, in a related work, proposed an efficient algorithm to train a model to sample from a product of energy-based distributions by minimizing a contrastive divergence, acknowledging that it is not generally straightforward to sample from a product of discrete distributions. More recently, \citet{diffusions} developed an MCMC-based algorithm to sample from a multiplicative composition of energy-base parameterized diffusion models, remarking the incorrectness of naive approaches based on sampling from the product of the reverse kernels. In the context of compounding GFlowNets with forward policies $(p^{(i)}_{F})_{1 \le i \le N}$ to sample from the corresponding product distribution $\prod_{1 \le i \le N} R_{i}$, however, one could exploit the structured nature of the model and naively attempt to use 
\begin{equation} \label{eq:app:p} 
    p_{F}(\cdot | s) \propto \prod_{1 \le i \le N} p_{F}^{(i)} (\cdot | s), 
\end{equation}
hoping that the resulting GFlowNet would sample from the correct product distribution. Notably, this approach fails due to the dependence of the normalizing constant of the preceding distribution on the state $s$ --- and it is unclear according to what distribution the sampled objects are distributed. \Cref{fig:app:pp} illustrates this for the problem of generating sequences: by combining the policies of the clients accordingly to \Cref{eq:app:p}, the culminating distribution drastically differs from the targeted one, even though the clients were almost perfectly trained. The question of which distributions one may obtain by composing the policies of expensively pre-trained GFlowNets is still open in the literature \cite{sculpting}. 

\paragraph{Implementing different training objectives for the clients.} \Cref{tab:tbtomato} suggests that the accuracy of EP-GFlowNet's distributional approximation is mostly independent of whether the clients implemented CB or TB as training objectives. Notably, the combination phase of our algorithm is designedly agnostic to how the local models were trained --- as long as they provide us with well-trained backward and forward policies. This is not constraining, however, since any practically useful training scheme for GFlowNets is explicitly based upon the estimation of such policies \cite{malkin2022trajectory, LingTrajectory, Foundations, robust}.

\begin{wrapfigure}[9]{r}{.2\textwidth} 
    \center 
    \vspace{-12pt} 
    \includegraphics[width=.8\linewidth]{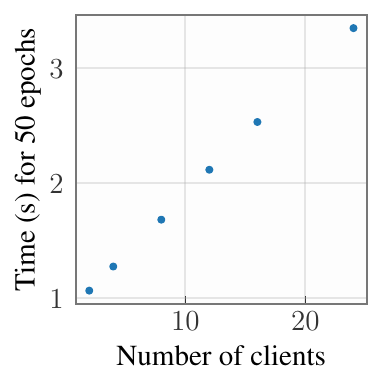}
    \caption{Aggregation phase's 50-epoch cost.} 
    \label{fig:epochs}
\end{wrapfigure}
\paragraph{Aggregation phase' computational cost.} \Cref{fig:epochs} shows that the per-epoch cost of training of a naively implemented EP-GFlowNets' aggregation phase, which we measure for the task of multiset generation, increases roughly linearly as a function of the number of worker nodes. Nonetheless, we remark that, in theory, one could achieve a sublinear relationship between the computational cost and the number of clients by parallelizing the local policies' evaluation when minimizing the aggregating balance loss function $\mathcal{L}_{AB}$, which is not done by our implementation.  

\begin{wrapfigure}{r}{.2\textwidth} 
    \includegraphics[width=.8\linewidth]{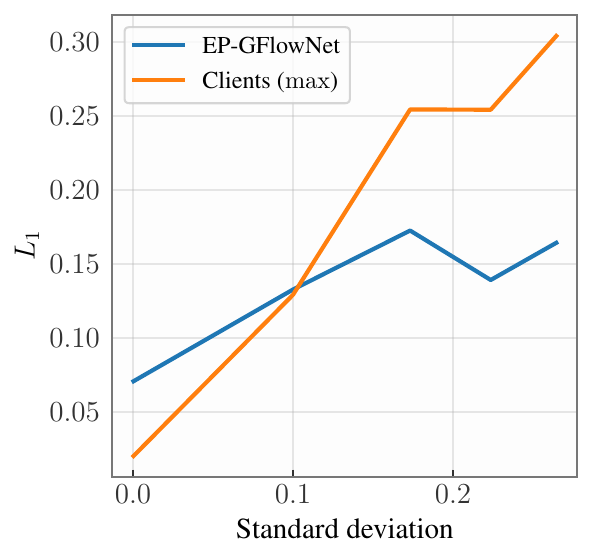}
    \caption{EP-GFlowNet's vs Clients' accuracy.}
    \label{fig:app:clients} 
\end{wrapfigure}
\noindent\textbf{Sensibility of EP-GFlowNet to imperfectly trained clients.} \Cref{thm:robustness} ensures that EP-GFlowNets can be trained to accurately sample from the combined target when the clients are sufficiently --- but maybe imperfectly --- learned. However, even when clients are imperfectly trained, EP-GFlowNets may achieve a relatively good approximation to the target. To exemplify this, we consider the task of multiset generation with $4$ clients trained on multiplicatively noisy targets, namely, $\log R'_{i}(x) = \log R_{i}(x) + \epsilon$ with $\epsilon \sim \mathcal{N}_{|\mathcal{X}|}(\mathbf{0}, \sigma^{2} \mathbf{I})$ sampled from a $|\mathcal{X}|$-dimensional normal distribution. Then, we perform EP-GFlowNets' aggregation phase and compare the learned distribution with the product $\prod_{1 \le i \le 4} R_{i}$. Importantly, \Cref{fig:app:clients} shows that the accuracy of the aggregated model is comparable and often better than that of the most inaccurate client for $\sigma^{2} \in \{0.000, 0.002, 0.004, 0.006, 0.008, 0.01\}$.  

\section{Federated Bayesian Network Structure Learning (BNSL)} 
\label{sec:app:dags} 

\begin{figure}
    \centering
    \includegraphics[width=\textwidth]{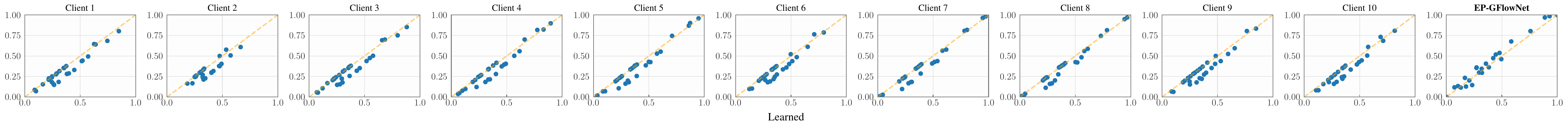}
    \caption{\textbf{Federated Bayesian network structure learning.} Each plot shows, for each pair of nodes $(U, V)$, the expected (vertical) and learned (horizontal) probabilities of $U$ and $V$ being connected by an edge, $\mathbb{P}[U \rightarrow V]$, and of existing a directed path between $U$ and $V$, $\mathbb{P}[U \rightsquigarrow V]$. Notably, EP-GFlowNet accurately matches the target distribution over such edges' features.}
    \label{fig:dags}
\end{figure}

Bayesian networks are often used to describe the statistical dependencies of a set of observed variables \cite{sem_ii, pearl18, pearl9, pearl1988probabilistic}; however, learning their structure from data is very challenging due to the combinatorially many possibilities. 
For example, when such variables represent some form of gene expression, we may want to infer the gene regulatory network describing the genes' causal relationships \cite{Husmeier2003}.  
Conventionally, structure learning is carried out centrally --- with all data gathered in a single place \cite{reisach2021beware, lorch2021dibs}. Nonetheless, such datasets are often small and insufficiently informative about the underlying structure. On the other hand, the rapid development of information technologies has significantly lowered the barrier for multiple parties (e.g., companies) to collaboratively train a model using their collected and privacy-sensitive datasets. In this context, we extend the work of \citet{deleu2022bayesian, deleu2023joint} and show that EP-GFlowNets can be efficiently used to learn a belief distribution over Bayesian networks in a federated setting \cite{Ng2022federated}.  

\begin{wrapfigure}[14]{r}{.45\textwidth} 
    \centering 
    \includegraphics[width=\linewidth]{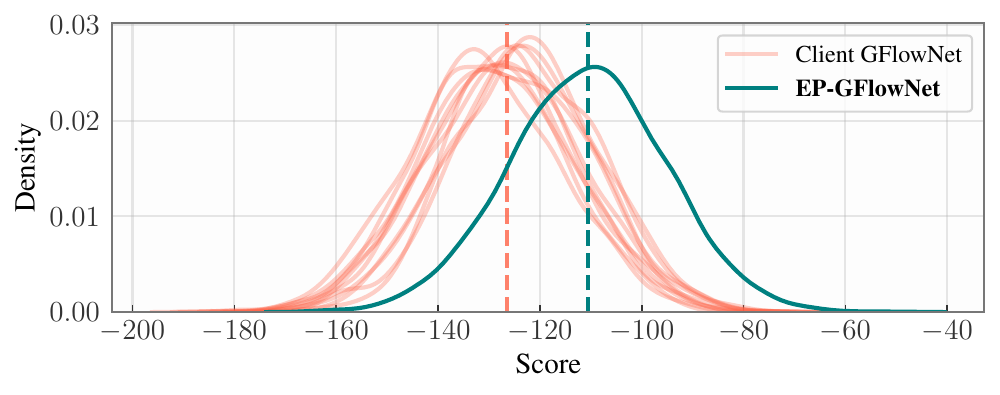} 
    \caption{The distribution learned in a federated manner (\textcolor{teal}{green}) finds significantly higher scoring graphs wrt the complete data than any of its local counterparts (\textcolor{red}{red}). The dashed lines represent the avg. scores for the global and local models.} 
    \label{fig:fed} 
\end{wrapfigure}
For this, let $\mathbf{X} \in \mathbb{R}^{d}$ be a random variable and $\mathbf{B} \in \mathbb{R}^{d \times d}$ be a sparse real-valued matrix with sparsity pattern determined by a directed acyclic graph $\mathcal{G}$ with nodes $\{1, \dots, d\}$. We assume that $\mathbf{X}$ follows a linear \textit{structural equation model} (SEM) 
\begin{equation} \label{eq:gaussian} 
    \mathbf{X} = \mathbf{B} \mathbf{X} + \mathbf{N}, 
\end{equation}
with $\mathbf{N} \sim \mathcal{N}(\mathbf{0}, \sigma^{2} I)$ representing independent Gaussian noise \cite{sem_ii, deleu2022bayesian}. Also, we consider $K$ fixed clients, each owning a private dataset $\mathcal{D}_{k} = \{\mathbf{x}_{1}, \dots, \mathbf{x}_{N}\}$, and our objective is to learn a belief distribution 
\begin{equation} \label{eq:dags} 
    R(G) = \prod_{1 \le k \le K} R_{k}(G) 
\end{equation}
over DAGs, in which $R_{k}(G) = p(\mathcal{D}_{k} | G) p(G)$ and $p(\mathcal{D}_{k} | G) = p(\mathcal{D}_{k} | \hat{\mathbf{B}}_{k}, G)$, with $\hat{\mathbf{B}}_{k}$ being client $k$'s maximum likelihood estimate of $\mathbf{B}$ given $G$ \cite{deleu2022bayesian}. Intuitively, the belief $R$ assigns high probability to DAGs which are probable under each of $R_{k}$'s and may be seem as a multiplicative composition of the local models \cite{sculpting, diffusions}. Notably, \citet{Ng2022federated} considered a similar setup, iteratively solving a $L_{1}$-regularized continuous relaxation of the structure learning problem to find a globally optimum DAG \cite{zheng2018dags}. However, in contrast to EP-GFlowNets, \citet{Ng2022federated}'s approach doesn't always guarantee that the obtained graph is acyclic \cite{geffner2022deep, zheng2018dags}; it doesn't provide a belief distribution over the DAGs, which could be successively refined by probing an expert \cite{bharti2022approximate, dasilva2023humanintheloop}; and it is not embarrassingly parallel, requiring many communication steps to be fulfilled.  

\noindent\textbf{Experimental setup.} To show that EP-GFlowNets can accurately sample from the belief distribution defined by \Cref{eq:dags} in a distributed setting, we first simulate a dataset of $400$ independent data points according to the SEM outlined in \Cref{eq:gaussian} with $d = 4$ variables. Next, we evenly partition this dataset between $4$ clients, which then train their own DAG-GFlowNets  \cite{deleu2022bayesian} based on their individual datasets for $10000$ epochs using AdamW \cite{loshchilov2018decoupled}. Finally, the locally trained DAG-GFlowNets are aggregated by minimizing the aggregating balance and the resulting model is compared to the belief distribution in \Cref{eq:dags}. See \Cref{sec:dags} for further discussion on our empirical observations for this domain.  

\noindent\textbf{Results.} \Cref{fig:dags} s that EP-GFlowNet adequately learns a distribution over DAGs in a federated setting. Correspondingly, \Cref{fig:fed} stresses that the collaboratively trained GFlowNet finds graphs with higher scores (graph-conditioned maximum log-likelihood) relatively to the private models. See \Cref{sec:dags} for discussion.

\section{Related work} 
\label{sec:app:work} 

\noindent\textbf{GFlowNets} were originally proposed as a reinforcement learning algorithm tailored to the search of diverse and highly valuable states within a given discrete environment \citep{Bengio2021}. Recently, these algorithms were successfully applied to the discovery of biological sequences \citep{sequence}, robust scheduling of operations in computation graphs \citep{robust}, Bayesian structure learning and causal discovery \citep{deleu2022bayesian, deleu2023joint, dasilva2023humanintheloop, dyngfn}, combinatorial optimization \citep{Zhang2023}, active learning \citep{multifidelity}, multi-objective optimization \citep{mogfn}, and discrete probabilistic modeling \citep{discretegfn_i, discretegfn_ii, discretegfn_iii}. \cite{Foundations} formulated the theoretical foundations of GFlowNets. Correlatively, \citep{theory} laid out the theory of GFlowNets defined on environments with a non-countable state space. \cite{stochastic} and \cite{quantiles} extended GFlowNets to environments with stochastic transitions and rewards. Concomitantly to these advances, there is a growing literature that aims to better understand and improve this class of algorithms \citep{markovchains, towards, malkin2023gflownets}, with an emphasis on the development of effective objectives and parametrizations to accelerate training convergence \citep{pan2023generative, LingTrajectory, malkin2022trajectory, deleu2022bayesian}. In recent work, for instance, \cite{LingTrajectory} proposed a novel residual parametrization of the state flows that achieved promising results in terms of speeding up the training convergence of GFlowNets. More specifically, the authors assumed the existence of a function $\mathcal{E} \colon \mathcal{S} \rightarrow \mathbb{R}$ such that (i) $\mathcal{E}(s_{o}) = 0$ and (ii) $\mathcal{E}(x) = - \log R(x)$ for each terminal state $x \in \mathcal{X}$ and reparameterized the state flows as $\log F(s, \phi_{S}) = - \mathcal{E}(s) + \log \tilde{F}(s, \phi_{S})$. This new training scheme was named \textit{forward looking (FL) GFlowNets} due to the inclusion of partially computed rewards in non-terminal transitions.    
Notably, both \citet{malkin2023gflownets}  and \citep{robust} proposed using the variance of the a TB-based estimate of the log partition function as a training objective based on the variance reduction method of \citet{vargrad}.
It is important to note one may use stochastic rewards \citep[see][]{Foundations, quantiles} for carrying out distributed inference, in the same fashion of, e.g., distributed stochastic-gradient MCMC \citep{el-mekkaoui_652, vono22}. 
Notably, stochastic rewards have also been used in the context of causal structure learning by \citet{deleu2023joint}.
However, it would require many communication steps between clients and server to achieve convergence --- which is one of the bottleneck EP-GFlowNets aim to avoid. 

\noindent\textbf{Distributed Bayesian inference} mainly concerns the task of approximating or sampling from a posterior distribution given that data shards are spread across different machines. This comprises both federated scenarios ~\citep{el-mekkaoui_652, vono22} or the ones in which we arbitrarily split data to speed up inference~\citep{Scott}. Within this realm, there is a notable family of algorithms under the label of \emph{embarrassingly parallel MCMC}~\citep{Neiswanger2014}, which employ a divide-and-conquer strategy to assess the posterior. These methods sample from subposteriors (defined on each user's data) in parallel, subsequently sending results to the server for aggregation. The usual approach is to use local samples to approximate the subposteriors with some tractable form and then  aggregate the approximations in a product. In this line, works vary mostly in the approximations employed. For instance, \citet{Mesquita2019} apply normalizing flows, \citep{Nemeth2018} model the subposteriors using Gaussian processes, and \citep{Wang+others:2015}  use hyper-histograms. It is important to note, however, that these works are mostly geared towards posteriors over continuous random variables.

\noindent\textbf{Federated learning} was originally motivated by the need to train machine learning models on privacy-sensitive data scattered across multiple mobile devices --- linked by an unreliable communication network~\citep{daneaistats2020}. While we are the first tackling FL of GFlowNets, there are  works on learning other generative models in federated/distributed settings, such as for generative adversarial networks~\citep{fgan1, fgan2, fgan3} and variational autoencoders~\citep{fvae}. Critically, EP-GFlowNets enable the collaborative training of a global model without disclosing the data underlying the clients' reward functions (defined by a neural network \cite{sequence} or a posterior distribution \cite{deleu2022bayesian, deleu2023joint, dasilva2023humanintheloop}, for instance); however, it does not inherently preserve the privacy of the reward functions themselves, which may be accurately estimated from their publicly shared policy networks. Nevertheless, a future work may investigate formally to which extent imperfectly trained policy networks ensure some form of differential privacy over the clients' rewards.  

\paragraph{Comparing EP-GFlowNets to single-round FedAVG \cite{mcmahan2017communication}} As EP-GFlowNets are the first method enabling embarrassingly parallel inference over discrete combinatorial spaces, there are no natural baselines to our experiments. For completeness, however, we use a single round of FedAVG \cite{mcmahan2017communication} by training local models (GFlowNets) until convergence and aggregating them in parameter space, assuming all policy networks share the same architecture. Note, however, that this baseline does not enjoy any guarantee of correctness and may lead to poor results. To validate this, we have run an experiment for the multiset generation task using FedAvg for aggregation. The resulting model attained an $L_{1}$ error of $1.32$, roughly two orders of magnitude worse compared to the $0.04$ of EP-GFlowNets.


\end{document}